%% file: heatmaps_ibcc.tex
\pdfoutput=1
\RequirePackage[fleqn]{amsmath}

\documentclass[a4paper]{llncs}
\usepackage[T1]{fontenc} 
\usepackage{graphicx}
\usepackage[fleqn]{amsmath}
\usepackage{amssymb}
\usepackage{amstext}
\usepackage{amsfonts}
\usepackage{cite}
\usepackage{algorithmic}
\usepackage{array}
\usepackage[caption=false,font=footnotesize]{subfig}
\usepackage{url}
\usepackage{times}
\usepackage[]{algorithm2e}
 
\newcommand{\bs}{\boldsymbol}  
\newcommand{\wrtd}{\mathrm{d}}
\title{ 
Bayesian Heatmaps: Probabilistic Classification with Multiple Unreliable Information Sources
}

\author{Edwin Simpson\inst{1\and 2} \and
Steven Reece\inst{2} \and Stephen J. Roberts\inst{2}}
\institute{Ubiquitous Knowledge Processing Lab,
Department of Computer Science,
Technische Universit{\"a}t Darmstadt, Germany.
\email{simpson@ukp.informatik.tu-darmstadt.de}
\and
Department of Engineering Science, 
University of Oxford, UK, 
\email{\{reece, sjrob\}@robots.ox.ac.uk} }

\begin{document}

\maketitle

% CUTTING DOWN (old notes, pre ICDM)

% PUT anything that is cut into a supplementary material paper that will act as an appendix.

% References that give clues about authors.

% Merging some of noise and bias. Summarise how similar the results are between noise and bias.

% Remove sparse reports plot?

% Unnecessary derivations -- to appendix/supplementary material.

%CUTS MADE (pre ICDM)

% cutting some explanation about length scales

% FOR LONG PAPER:

% Flesh out the discussion of learning length scale from other datasets. -- could cut from conference paper. Take out the ML bit. More detail about the assumptions made here.

% for long paper: shorten "the spread of emergencies across an earthquake area is similar to the distrubtion of building damage".

\begin{abstract}

% Abstract:

Unstructured data from diverse sources, such as social media and aerial imagery, can provide 
valuable up-to-date information for intelligent situation assessment.
Mining these different information sources could bring major benefits to applications such as situation awareness in disaster zones and mapping the spread of diseases.
Such applications depend on classifying the situation across a region of interest, 
which can be depicted as a spatial ``heatmap". 
Annotating unstructured data using crowdsourcing or automated classifiers produces 
individual classifications at sparse locations that typically contain many errors.
We propose a novel Bayesian approach that models the relevance, error rates and bias of each information source, enabling us to learn a spatial Gaussian Process classifier by aggregating data from multiple sources with varying reliability and relevance. 
Our method does not require gold-labelled data and can make predictions at any location in an area of interest given only sparse observations.
We show empirically that our approach can handle noisy and biased data sources, 
and that simultaneously inferring reliability and transferring information between neighbouring reports leads to more accurate predictions.
We demonstrate our method on two real-world problems from disaster response, 
showing how our approach reduces the amount of crowdsourced data required
and can be used to generate valuable heatmap visualisations from SMS messages and satellite images.

 \end{abstract}

% For peer review papers, you can put extra information on the cover
% page as needed:
% \ifCLASSOPTIONpeerreview
% \begin{center} \bfseries EDICS Category: 3-BBND \end{center}
% \fi
%
% For peerreview papers, this IEEEtran command inserts a page break and
% creates the second title. It will be ignored for other modes.
%\IEEEpeerreviewmaketitle

%%%%%%%%%%%%%%%%%%%%%%%%%%%%%%%%%%%%%%%%%%%%%%%%%%%%%%%%%%%%%%%%%

\input{sections/intro}
\input{sections/method}
\input{sections/experiments}
\input{sections/discussion}

%%%%%%%%%%%%%%%%%%%%%%%%%%%%%%%%%%%%%%%%%%%%%%%%%%%%%%%%%%%%%%%%%%%%%%%%%%%%%%%%

% use section* for acknowledgment
\section*{Acknowledgments}

We thank Brooke Simmons at Planetary Response Network for invaluable support and data.
This work was funded by EPSRC ORCHID programme grant (EP/I011587/1). 

% \addcontentsline{toc}{chapter}{Bibliography}
%\bibliographystyle{apalike}
%\bibliographystyle{IEEEtran}
\bibliographystyle{splncs03}
\bibliography{simpson}

\end{document}

%% file: sections/intro.tex
\section{Introduction\label{sec:intro}}

Social media enables members of the public to post real-time text
messages, videos and photographs describing events taking place close
to them.
While many posts may be extraneous or misleading, 
social media nonetheless provides streams of up-to-date information across a wide area.
% when trusted sources are limited. 
For example, after the Haiti 2010 earthquake, Ushahidi gathered thousands of text messages that provided valuable first-hand information about the disaster situation \cite{morrow2011independent}.  
An effective way to extract information from large unstructured datasets such as these 
is to employ crowds of non-expert annotators, 
as demonstrated by Galaxy Zoo\cite{lintott2008galaxy}.
Besides social media, crowdsourcing provides a means to obtain geo-tagged annotations from other unstructured data sources such as imagery from satellites or unmanned aerial vehicles (UAV).

In scenarios such as disaster response, we wish 
to infer the situation across a region of interest by combining
 annotations from multiple information sources.
For example, we may wish to determine which areas are currently flooded, 
the level of damage to buildings in an earthquake zone, 
or the type of terrain in a specific area
from a combination of SMS reports and satellite imagery.
The situation across an area of interest can be visualised using a \emph{heatmap} 
(e.g. Google Maps heatmap layer\footnote{https://developers.google.com/maps/documentation/javascript/examples/layer-heatmap}), 
which overlays colours onto a map to indicate
the intensity or probability of phenomena of interest. 
Probabilistic methods have been used to generate heatmaps from 
observations at sparse, point locations\cite{kottas2007bayesian,adams2009tractable,samo2014scalable},
using a Bayesian treatment of Poisson process models.
However, these approaches model the rate of occurrence of events, 
so are not suitable for classification problems.
Instead, a Gaussian process (GP) classifier can be used to model
a class label that varies smoothly over space or time.
This uses a latent function over input coordinates, which is mapped
through a sigmoid function to obtain probabilities\cite{rasmussen_gaussian_2006}. 
However, standard GP classifiers are unsuitable for heterogeneous, crowdsourced data since they do not account for the differing relevance, error rates and bias of individual information sources and annotators.

A key challenge in exploiting crowdsourced information is to account for its unreliability and combine it with trusted data as it becomes available, such as reports from experienced first responders in a disaster zone. 
For regression problems, differing levels of accuracy can be handled using 
sensor fusion approaches such as \cite{meng2015truth, venanzi2013}. The approach of 
\cite{venanzi2013} uses heteroskedastic GPs to produce heatmaps that account for sensor accuracy 
through variance scaling. 
This method could be applied to spatial classification by mapping GPs through a softmax function.
However, such an approach cannot handle label bias or accuracy that depends on the true class.
Recently, \cite{Long2016}, proposed learning a GP classifier from
crowdsourced annotations, but their method uses a coin-flipping noise model 
that would suffer from the same drawbacks as adapting \cite{venanzi2013}.
Furthermore they train the model using a maximum likelihood (ML) approach, 
which may incorrectly estimate reliability when data for some workers is insufficient\cite{simpsonlong,raykar12,kim2003}.

For classification problems, 
each information source can be modelled by a confusion matrix \cite{dawid_maximum_1979},
which quantifies the likelihood of observing a particular annotation from an information source given the true class label. This approach naturally accounts for bias toward a particular answer and varying accuracy depending on the true class, and has been 
shown to outperform techniques such as majority voting and weighted sums\cite{simpsonlong,raykar12,kim2003}.
Recent extensions following the Bayesian treatment of \cite{kim2003} can further improve results:
 by identifying clusters of crowd workers with shared confusion matrices 
\cite{moreno_bayesian_2015,venanzi2014community};
accounting for the time each worker takes to complete a task\cite{venanzi2016time};
additionally modelling language features in text classification tasks\cite{Felt2016SemanticAA, simpson2015language}.
However, these methods depend on receiving multiple labels from different workers for the same data points,
or, in the case of \cite{Felt2016SemanticAA, simpson2015language}, on correlations between text features and target classes.
None of the existing confusion matrix-based approaches can model the spatial 
distribution of each class,
and therefore, when reports are sparsely distributed over an area of interest,
they cannot compensate for the lack of data at each location.

In this paper, we propose a novel Bayesian approach to aggregating sparse, geo-tagged reports 
from sources of varying reliability, 
which combines independent Bayesian classifier combination (IBCC) \cite{kim2003} 
with a GP classifier to infer
 discrete state values across an area of interest.
Our model, \emph{HeatmapBCC}, assumes that states at neighbouring locations are correlated, allowing us to fuse neighbouring reports and interpolate between them to predict the state at locations with no reports.
HeatmapBCC uses confusion matrices to model the error rates, relevance and bias of each
 information source, permitting the use of non-expert crowds providing heterogeneous annotations.
 The GP handles the uncertainty that arises from sparse spatial data in a principled Bayesian manner, 
 allowing us to incorporate prior information, 
 such as physical models of disaster events such as earthquakes, 
 and visualise the resulting posterior distribution as a spatial heatmap.
We derive a variational inference method that is able to learn the 
reliability model for each information source
without the need for ground truth training data. 
This method learns full distributions over latent variables that can be used 
to prioritise locations for further data gathering using an active learning approach.
The next section presents in detail the HeatmapBCC model, and provides details of our efficient approximate inference algorithm. The following section then provides an empirical evaluation of our method on both synthetic and real-world problems, showing that HeatmapBCC can outperform rival methods.
We make our code publicly available at \url{https://github.com/OxfordML/heatmap_expts}.

%\subsection{Authors' notes}
%
%Remove this section later.
%
%Possibility of splitting into two papers for different audiences:
%
%1. Statistical methodology paper for NIPS: Accounting for trust when
%training a GP classifier. Talk about this as a general method for learning a GP
%classifier from untrusted sources. Experiments run on Ushahidi, Time series
%(Abby's data?)...
%
%2. Bayesian heatmaps: fusing reports from a heterogeneous crowd. Aimed at
%Crowdsourcing community (HCOMP? AAAI? ICWSM?). Key insight: we can
%construct heatmaps by treating messages posted by crowd as weak classifications.
%Show Ushahidi + another social media dataset. Test with both Ushahidi categories
%as labels and text --> extensive labelling is not required to construct the
%heatmaps. We can construct them efficiently by selecting reports iteratively,
%looking in areas where map is uncertain (active learning). Problem: paper 1
%will have introduced the application to crowdsourcing, as may the AAMAS paper.
%Novelty lies in full system description and detailed study of crowdsourcing
%application.
%
%\subsection{Related Work}
%
%Spatial cox processes learn the intensity of reports. This doesn't usually
%account for different reliability of reporters, and outputs a report intensity
%as a prediction. The latter is not useful if you want to learn damage
%probability in regions with high mobile users. Advantage is it gives you a
%non-binary prediction such as damage intensity.

%% file: sections/method.tex
\section{The HeatmapBCC Model\label{sec:model}}

Our goal is to classify locations of interest, e.g. to identify 
them as ``flooded'' or ``not flooded''. We can then choose
locations in a grid over an area of interest and 
plot the classifications on a map as a spatial \emph{heatmap}.
The task is to infer a vector $\bs t^*\in\{1,..,J\}^{N^*}$
of target state values at $N^*$ locations $\bs X^*$, where $J$ is the number of state values or classes.
Each row $\bs x_i$ of matrix $\bs X^*$ is a coordinate vector that specifies a point on the map. 
We observe a matrix of potentially unreliable geo-tagged \emph{reports},
$\bs c\in\{1,..,L\}^{N\times S}$,
with $L$ possible discrete values, from $S$ 
different information sources at $N$ training locations $\bs X$.
%The locations in
%$\bs X$ may overlap with the set of output grid points in
%$\bs X^*$, but we typically expect the input
%points to be sparse, in contrast to the evenly-spaced output grid used to
%produce a heatmap.  
%In practice, the matrix $\bs c$ is also likely as not many sources or crowd members will
%supply data for every location.

HeatmapBCC assumes that each report label $c_{i}^{(s)}$, from information source $s$, 
at location $\bs x_i$,
is drawn from $ c_{i}^{(s)} | t_{i}, \bs\pi^{(s)} 
 \sim \mathrm{Categorical}(\bs\pi_{t_{i}}^{(s)} )$.
 The target state, $t_{i}$, selects the row, $\bs\pi_{t_{i}}^{(s)}$,
of a \emph{confusion matrix}\cite{dawid_maximum_1979,simpsonlong}, 
$\bs\pi^{(s)}$, which describes the errors and biases of $s$ 
as a dependency between the report labels and
the ground truth state, $t_{i}$.
As per standard IBCC \cite{kim2003}, 
the reports from each information source are 
conditionally independent of one another given target $t_i$,
and each row of the confusion matrix is drawn from 
$ \bs\pi_j^{(s)} | \bs\alpha_{0,j}^{(s)} 
\sim \mathrm{Dirichlet}(\bs\alpha_{0,j}^{(s)})$. 
The hyperparameters $\bs\alpha_{0,j}^{(s)}$ encode the prior trust in $s$.

%we assume that at each location there is a latent function value f, which
% we can map to a state probability rho using a function sigma.

We assume that state $t_{i}$ at location $\bs x_i$ is drawn from 
a categorical distribution,
$ t_{i} | \bs\rho_{i} \sim \mathrm{Categorical}(\bs\rho_{i})$,
where $\rho_{i,j} = p(t_{i}=j | \bs\rho_{i} ) \in[0,1]$ is the probability
of state $j$ at location $\bs x_i$.
The generative process for state probabilities, $\bs\rho$, is as follows.
First, draw latent functions for classes $j\in\{1,..,J\}$ from a Gaussian process prior:
$f_j \sim \mathcal{GP}(m_j, k_{j, \bs\theta}/\varsigma_j)$, 
where $m_j$ is the prior mean function, $k_j$ is the prior covariance
function, $\bs\theta$ are hyperparameters of the covariance function, 
and $\varsigma_j$ is the inverse scale.
Map latent function values $f_j(\bs x_i)\in\mathcal{R}$ to state probabilities: 
$\bs\rho_{i} = \sigma(f_1(\bs x_i),..,f_J(\bs x_i))\in[0,1]^J$.
Appropriate functions for $\sigma$ include the logistic sigmoid and probit functions
for binary classification, and softmax and multinomial probit for multi-class classification.
We assume that $\varsigma_j$ is drawn from a conjugate gamma hyperprior, 
$\varsigma_j \sim \mathcal{G}\left( a_{0}, b_{0} \right)$,
where $a_{0}$ is a shape parameter and $b_{0}$ is the inverse scale.

While the reports, $c_{i}^{(s)}$, are modelled in the same way as standard IBCC
\cite{kim2003}, HeatmapBCC introduces a location-specific state probability,
$\bs\rho_{i}$, to replace the global class proportions,
$\bs\kappa$, which IBCC \cite{simpsonlong}
assumes are constant for all locations.
Using a Gaussian process prior means the state probability varies reasonably
smoothly between locations, thereby encoding correlations in the distribution
over states at neighbouring locations. 
The covariance function is chosen to suit the scenario we wish to model 
and may be tailored to specific spatial phenomena (the geo-spatial impact of an earthquake, for example). 
The hyperparameters, $\bs\theta$, typically include
a length-scale, $l$, which controls the smoothness of the function. 
Here, we assume a stationary covariance function of the form 
$  k_{j,\bs\theta}\left(\bs x, \bs x'\right) = k_j \left(|\bs x - \bs x'|, l\right)$,
where $k$ is a function of the distance between two points and the length-scale, $l$.
The joint distribution for the complete model is:
\begin{eqnarray*}
 p\left(\bs c, \bs t, \bs f_1, .., \bs f_J, \bs \varsigma_1,.., \bs\varsigma_J, \bs\pi^{(1)} \!,...,\bs\pi^{(S)} |
 \bs\mu_1,..,\bs\mu_J, \bs K_1,..,\bs K_J, \bs\alpha_0^{(1)} \! ,..,\bs\alpha_0^{(S)}\right) = & & 
\nonumber \\
\prod_{i=1}^N \!\left\{ \rho_{i,t_i} \!
%{\bs x\in \bs X} \!\left\{ \rho_{t_{\bs x},\bs x} \!
 \prod_{s=1}^S%\prod_{s \in R_{\bs x}} 
% \pi^{(s)}_{t_{\bs x},c_{\bs x}^{(s)}} \!\right\}
 \pi^{(s)}_{t_i,c_i^{(s)}} \!\right\}
\prod_{j=1}^J \left\{
p\left( \bs f_j | \bs\mu_j, \bs K_j/\varsigma_j \right) 
p\left( \varsigma_j | a_0, b_0 \right)
 \prod_{s=1}^S p\left( \bs \pi^{(s)}_j | \bs \alpha^{(s)}_{0,j}\right) \right\},
%\label{eq:joint}
&&
\end{eqnarray*}
where $\bs f_j=\left[ f_j(\bs x_1), .., f_j(\bs x_N) \right]$,
$\bs\mu_j = \left[ m_j(\bs x_1),.., m_j(\bs x_N) \right]$, 
and  
$\bs K_j\in \mathbb{R}^{N\times N}$ with elements $K_{j,n, {n'}}=k_{j,\bs\theta}(\bs x_n,\bs x_{n'})$.

\section{Variational Inference for HeatmapBCC}\label{sec:vb}

We use \emph{variational Bayes (VB)} to efficiently approximate the posterior distribution 
over all latent variables,
allowing us to handle streaming data reports online by restarting the VB algorithm from 
the previous estimate as new reports are received. 
To apply variational inference, we replace the exact posterior distribution with a variational approximation that factorises into separate latent variables and parameters:
\begin{eqnarray}
p(\bs t, \bs f, \bs\varsigma, \bs \pi^{(1)},..,\bs \pi^{(S)} | \bs c, \bs\mu, \bs K, \bs\alpha_0^{(1)},..,\bs\alpha_0^{(S)}) 
\approx q(\bs t) \prod_{j=1}^J \left\{ q(\bs f_j)q(\varsigma_j)\prod_{s=1}^S q\left(\bs\pi_j^{(s)}\right) \right\}. &&
\nonumber
\end{eqnarray}
% where $\bs Z=\{\bs t, \bs f, \bs\varsigma, \bs \pi^{(1)},..,\bs \pi^{(S)}\}$ is the set of
% all latent variables and parameters.
We perform approximate inference by optimising the variational posterior using
Algorithm \ref{alg:vbheatmapbcc}. In the remainder of this section we define the variational factors $q()$, expectation terms, variational lower bound and prediction step required by the algorithm. 
%by iteratively updating the log of each variational factor, $\log q(z_i)$, in turn, 
% for each set of latent variables, $z_i\in\bs Z$, until it converges to an approximate solution. 
\begin{algorithm}
\DontPrintSemicolon 
\SetKwInOut{Input}{input}\SetKwInOut{Output}{output}
\Input{Hyperparameters $\bs\alpha_0^{(s)}\ \forall s$, $\bs\mu_j\ \forall j$, $\bs K$, $a_0$, $b_0$;
observed report data $\bs c$}
{ \label{step:1} Initialise $q\left(\bs f_j\right)\ \forall j$, $q\left(\bs\pi_j^{(s)}\right)\ \forall j\ \forall s$, and
$q(\varsigma_j)\forall j$ randomly\; }
\While{variational lower bound not converged}{
\label{step:2} Calculate $\mathbb{E}\left[\log \bs\rho \right]$ 
and $\mathbb{E}\left[\log\bs\pi^{(s)} \right], \forall s$ given current factors 
$q\left(\bs f_j\right)$ and $q\left(\bs\pi_j^{(s)}\right)$ \;
Update $q(\bs t)$ given $\mathbb{E}\left[\log\bs\pi^{(s)} \right], \forall s$ and $\mathbb{E}\left[\log \bs\rho \right]$ \;
Update $q\left(\bs\pi_j^{(s)}\right), \forall j, \forall s$ given current estimate for $q(\bs t)$ \;
\label{step:4} Update $q\left(\bs f_j\right), \forall j$ current estimates for $q(\bs t)$ and $q(\varsigma_j), \forall j$ \;
\label{step:varsigma} Update $q(\varsigma_j),\forall j$ given current estimate for $q\left(\bs f_j\right)$ \;
}
\Output{ \label{step:7} Use converged estimates to predict $\bs\rho^*$ and $\bs t^*$ at output points $X^*$}
\;
\caption{
VB algorithm for HeatmapBCC}\label{alg:vbheatmapbcc}
\end{algorithm}

\paragraph{\bf Variational Factor for Targets, $\bs t$:}
\begin{flalign}
& \log q(\bs t) = 
 \sum_{i=1}^N \left\lbrace \mathbb{E}[\log \rho_{i,t_{i}}] + \sum_{s=1}^S 
\mathbb{E}\left[\log\pi^{(s)}_{t_{i},c_{i}^{(s)}}\right]
\right\rbrace 
 + \mathrm{const}. &&
\end{flalign}
The variational factor $q(\bs t)$ further factorises into individual data
points, since the target value, $t_i$, at each input point, $\bs x_i$,
 is independent given the state probability vector 
 $\bs \rho_i$, giving $r_{i,j} := q(t_{i}=j)$ where $q(t_{i}=j) = q(t_{i}=j,\bs c_{i})/
\sum_{\iota\in J}q(t_{i}=\iota,\bs c_{i})$ 
and:
\begin{flalign}
q(t_{i}=j,\bs c_{i}) = \exp\left(
\mathbb{E}\left[\log \rho_{i,j}\right]
+ \sum_{s=1}^S
\mathbb{E}\left[\log\pi^{(s)}_{j,c_{i}^{(s)}}\right] \right). & &
 \label{eq:responsibilities}
\end{flalign}
Missing reports in $\bs c$ can be handled simply by omitting the term 
$\mathbb{E}\left[\log\pi^{(s)}_{j,c_{i}^{(s)}}\right]$
for information sources, $s$, that have not provided a report $c^{(s)}_i$.

\paragraph{\bf Variational Factor for Confusion Matrix Rows, $\bs \pi_j^{(s)}$:}
\begin{flalign*}
& \log q\left(\bs \pi_j^{(s)}\right) = \mathbb{E}_{\bs t}\left[\log p\left(\bs \pi^{(s)}|\bs 
t,\bs c\right)\right] 
 = \sum_{l=1}^L N^{(s)}_{j,l}\log\pi^{(s)}_{j,l} +
 \log p\left(\bs\pi^{(s)}_j|\bs\alpha_{0,j}^{(s)}\right) + \mathrm{const}. 
 , &&
\end{flalign*}
where $N^{(s)}_{j,l}= \sum_{i=1}^N r_{i,j}\delta_{l,c_{i}^{(s)}}$ are pseudo-counts
and $\delta$ is the Kronecker delta.
Since we assumed a Dirichlet prior, the variational distribution is also a Dirichlet,
$q(\bs\pi^{(s)}_j) = \mathcal{D}(\bs\pi^{(s)}_j | \bs\alpha_{j}^{(s)})$,
 with parameters
$\bs\alpha_{j}^{(s)} = \bs\alpha_{0,j}^{(s)}+\bs N^{(s)}_j$, 
where $\bs N^{(s)}_{j} =\left\{N^{(s)}_{j,l} | l \in[1,..,L] \right\}$.
Using the digamma function, $\Psi()$, 
the expectation required for Equation \ref{eq:responsibilities} is therefore:
\begin{flalign}
&
\mathbb{E}\left[\log\pi^{(s)}_{j,l}\right] = \Psi\left(\alpha_{j, l}^{(s)}\right) - \Psi\left(\sum_{\iota=1}^L \alpha_{j, \iota}^{(s)}\right).&&
\end{flalign}

%******* GP STUFF **************************************************************

\paragraph{\bf Variational Factor for Latent Function:}

%For steps \ref{step:2} and \ref{step:4} in the VB algorithm, we introduce new
%variational update equations that enable us to 
%track correlations between $\bs\rho_{i}$ at different locations
%by incorporating a Gaussian process over a latent function $f$, where $\rho_{i,j} = \sigma(\bs f)_{j,i}$. 
%We therefore perform updates in terms of the matrix of latent function values, $\bs f$, rather than $\bs \rho$ directly.
The variational factor $q(\bs f)$ factorises between target classes, since $t_i$ at each point is independent given $\bs\rho$. Using the fact that
$\mathbb{E}_{t_i}[\log\mathrm{Categorical}([t_i=j] | \rho_{i,j})] = r_{i,j}\log\sigma(\bs f)_{j,i}$, the factor 
for each class is:
\begin{flalign} 
& \log q(\bs f_j) = \sum_{i=1}^N r_{i,j}
\log\sigma(\bs f)_{j,i} + \mathbb{E}_{\varsigma_j}\left[  \log \mathcal{N}(\bs f_{j} | \bs\mu_{j}, \bs K_{j}/\varsigma_j) \right] + \mathrm{const}.  && \label{eq:qrho}
\end{flalign}
This variational factor cannot be computed analytically, 
but can itself be approximated using a variational method
based on the extended Kalman filter (EKF) \cite{reece2011determining, steinberg2014extended}
%can be used to infer the Gaussian process latent function $f_j$, and
%method is referred to as the extended Gaussian process (EGP) in \cite{steinberg2014extended}, in which the authors show that this is a variational inference method. Therefore, the EGP 
that is amenable to inclusion in our overall VB algorithm. 
%Since a single step of the EGP algorithm will increase the lower bound on the log marginal likelihood, we do not need to run the EGP update equation until convergence on every iteration of the main VB algorithm. This reduces the computational cost of early iterations of our main VB algorithm, when $q(\bs f_j)$ is likely to change significantly between iterations and refinement of the estimates is therefore unnecessary. 
Here, we present a multi-class variant of this method that applies ideas from \cite{girolami2006variational}.
We approximate the likelihood 
$p(t_{i}=j | \bs\rho_{i,j}) = \bs\rho_{i,j}$ with a Gaussian distribution, using 
$\mathbb{E}[\log \mathcal{N}([t_i=j] | \sigma( \bs f)_{j,i}, v_{i,j})] = \log \mathcal{N}\left(r_{i,j} | \sigma( \bs f)_{j,i}, v_{i,j}\right)$ 
to replace Equation \ref{eq:qrho} with the following:
\begin{flalign}
& \log q(\bs f_j) 
\approx \! \sum_{i=1}^N 
 \! \log \mathcal{N}(r_{i,j} | \sigma( \bs f)_{j,i}, v_{i,j})
 \!+ \mathbb{E}_{\varsigma_j}[ \log \mathcal{N} \left( \bs f_{ j} | \bs\mu_j, \bs K_j/\varsigma_j \right)]
 \!+ \! \mathrm{const},&&
\label{eq:rhopostapprox}
\end{flalign}
where $v_{i,j}=\rho_{i,j}(1 - \rho_{i,j})$ is the variance of the binary indicator variable $[t_i=j]$ given by the Bernoulli distribution. 
We approximate Equation \ref{eq:rhopostapprox} by linearising $\sigma()$ using a Taylor series
expansion to obtain a multivariate Gaussian 
distribution $q(\bs f_j) \approx \mathcal{N}\left(\bs f_j| \hat{\bs f}_{j},\bs \Sigma_j\right)$.  
Consequently, we estimate $q\left(\bs f_j\right)$ using EKF-like
equations\cite{reece2011determining,steinberg2014extended}:
\begin{flalign}
 \hat{\bs f}_{j} &= \bs\mu_{j} + \bs W 
                    \left( \bs r_{.,j} - \sigma(\hat{\bs f})_{j} + \bs G (\hat{\bs f}_{j} - \bs\mu_{j})
                    \right) \label{eq:hatf} \\ %\frac{}{\sqrt{\varsigma}}
\bs \Sigma_j &= \hat{\bs K}_j  - \bs W \bs G_j \hat{\bs K}_j  \label{eq:sigmaf}
\end{flalign}
where $\hat{\bs K}_j^{-1}  = \bs K_j^{-1}\mathbb{E}[\varsigma_j]$ 
and $\bs W=\hat{\bs K}_j  \bs G_j^T \left(\bs G_j \hat{\bs K}_j  \bs G_j^T + \bs Q_j\right)^{-1}$ is the Kalman gain, 
$\bs r_{.,j}=\left[r_{1,j},r_{N,j}\right]$ is the vector of probabilities of target state
$j$ computed using Equation \ref{eq:responsibilities} for the input points,
$\bs G_j \in \mathbb{R}^{N \times N}$ is the diagonal sigmoid Jacobian matrix 
and $\bs Q_j\in\mathbb{R}^{N\times N}$ is a diagonal observation noise variance
matrix. The diagonal elements of $\bs G$ are 
$\bs G_{j, i, i} = \sigma(\hat{\bs f}_{.,i})_j (1 - \sigma(\hat{\bs f}_{.,i})_j )$,
where $\hat{\bs f}=\left[\hat{\bs f}_1,..,\hat{\bs f}_J\right]$ is the matrix of mean values for all classes.

The diagonal elements of the noise covariance matrix are 
$Q_{j,i,i} = v_{i,j}$, which we approximate as follows.
Since the observations are Bernoulli distributed with an uncertain
parameter $\rho_{i,j}$, the conjugate prior over 
$\rho_{i,j}$ is a beta distribution with parameters 
$\sum_{j'=1}^J \nu_{0,j'}$ and $\nu_{0,j}$.
This can be updated to a posterior Beta distribution $p\left(\rho_{i,j} | r_{i,j}, \bs \nu_{0} \right) = \mathcal{B} \left( \rho_{i,j} | \nu_{\neg j}, \nu_j  \right)$,
where $\nu_{\neg j} = \sum_{j'=1}^J \nu_{0,j'} - \nu_{0,j} + 1 - r_{i,j}$ and $\nu_j = \nu_{0,j} + r_{i,j}$. 
We now estimate 
the expected variance: 
\begin{flalign}
 & v_{i,j}\approx \hat{v}_{i,j} = \int \left(\rho_{i,j} - \rho_{i,j}^2\right) \mathcal{B}\left( \rho_{i,j} | \nu_{\neg j}, \nu_{j}\right) \wrtd \rho_{i,j} = \mathbb{E}[\rho_{i,j}] - \mathbb{E}\left[\rho_{i,j}^2\right] 
\end{flalign}
\begin{flalign}
& \mathbb{E}[\rho_{i,j}]  = \frac{\nu_j} {\nu_j + \nu_{\neg j}}  && \label{eq:expec_post_beta} 
& \mathbb{E}\left[\rho_{i,j}^2\right]  = \mathbb{E}[\rho_{i,j}]^2 + \frac{\nu_j \nu_{\neg j} } {(\nu_j + \nu_{\neg j})^2(\nu_j + \nu_{\neg j} + 1)} . 
\end{flalign}
We determine values for the prior beta parameters, $\nu_{0,j}$, by moment matching with the 
prior mean $\hat{\rho}_{i,j}$ and variance $u_{i,j}$ of $\rho_{i,j}$, found using numerical integration.
According to Jensen's inequality, the convex function $\varphi(\bs Q) = \left(\bs G_j \bs K_j \bs G_j^T + \bs Q\right)^{-1}$ is a lower bound on 
$\mathbb{E}[\varphi(\bs Q)] = \mathbb{E}\left[(\bs G_j \bs K_j \bs G_j^T + \bs Q)^{-1}\right]$.  Thus our approximation provides a tractable estimate of the expected value of $\bs W$. 

The calculation of $\bs G_j$ requires evaluating the latent function at the input points $\hat{\bs f}_j$.  Further, Equation \ref{eq:hatf} requires $\bs G_j$ to approximate $\hat{\bs f}_j$, causing a
circular dependency.  Although we can fold our expressions for $\bs G_j$ and $\hat{\bs f}_j$ directly into the VB cycle and update each variable in turn, we found solving for $\bs G_j$ and $\hat{\bs
  f}_j$ each VB iteration facilitated faster inference.  We use the following iterative procedure to estimate $\bs G_j$ and $\hat{\bs f}_j$:
\begin{enumerate}
\item Initialise $\sigma(\hat{\bs f}_{.,i}) \approx \mathbb{E}[\bs\rho_{i}]$ using Equation \ref{eq:expec_post_beta}.
\item Estimate $\bs G_j$ using the current estimate of $\sigma(\hat{f}_{j,i})$.
\item Update the mean $\hat{\bs f}_{j}$ using Equation \ref{eq:hatf}, inserting the current estimate of $\bs G$. 
\item Repeat from step 2 until $\hat{\bs f}_{j}$ and $\bs G_j$ converge.
\end{enumerate}
The latent means, $\hat{\bs f}$, are then used to estimate the terms $\log \rho_{i,j}$
for Equation \ref{eq:responsibilities}:
\begin{flalign}
& \mathbb{E}[ \log \rho_{i,j}] = \hat{f}_{j,i} - \mathbb{E}\left[\log\sum_{j'=1}^J \exp(f_{j', i}) \right]. && \label{eq:elogrho|_softmax}
\end{flalign}
When inserted into Equation \ref{eq:responsibilities},
the second term in Equation \ref{eq:elogrho|_softmax} 
cancels with the denominator, so need not be computed.
% Therefore, we can use $\hat{\bs f}$ to directly compute our updates for step \ref{step:2}. 
%This completes the calculations relating to $q(\bs\rho)$ that are required by
%the VB algorithm for steps \ref{step:1} to \ref{step:4}.

\paragraph{\bf Variational Factor for Inverse Function Scale:}

%See RD Turner 2011 thesis
The inverse covariance scale, $\varsigma_j$, can also be inferred using VB by taking expectations
with respect to $\bs f$:
\begin{flalign*}
& \log q\left(\varsigma_j\right) = \mathbb{E}_{\bs \rho}[\log p(\varsigma_j | \bs f_j)] 
%&= \mathbb{E}_{\bs \rho}[\log p (\rho_{j,i} | m, \varsigma_j, \bs\theta)] + \log p(\varsigma_j |  a_{0}, b_{0} ) 
%- \log Z_{\varsigma_j}\\
=  \mathbb{E}_{\bs f_j}[\log\mathcal{N}(\bs f_{j} | \mu_{i}, \bs K_j/\varsigma_j)]
 + \log p(\varsigma_j |  a_{0}, b_{0} ) + \mathrm{const} &&
\end{flalign*}
which is a gamma distribution with shape $a=a_0 + \frac{N}{2}$ 
and inverse scale $b = b_0 + \frac{1}{2} \mathrm{Tr}\left(\bs K_j^{-1}
\left( \bs\Sigma_j + \hat{\bs f}_j \hat{\bs f}_j^T - 2 \bs\mu_{j,i}\hat{\bs f}_j^T - \bs\mu_{j,i}\bs\mu_{j,i}^T \right)\right)$.
%where $\mathbb{E}_{\bs f_j}[\bs f_j \bs f_j^T] = \Sigma_j + \hat{\bs f}_j \hat{\bs f}_j^T$ because E(x^2) = V(x) + E(x)^2
%where $\bs K' = \varsigma \bs K_j$ is the unscaled covariance matrix, which can be pre-computed once at the start of the algorithm
%as it does not depend on any of the variables that will be updated during the variational inference process. 
We use these parameters to compute the expected latent model precision, $\mathbb{E}[\varsigma_j] = a/b$ in Equation~\ref{eq:sigmaf}, and for the lower bound described in the next section we also require 
$\mathbb{E}_q[\log(\varsigma_j)] = \Psi(a) - \log(b)$. 

%and for the lower bound in the next subsection, $\mathbb{E}[\log\varsigma_j] = \Psi(a) - \log(b)$.

\paragraph{\bf Variational Lower Bound:}
%In step \ref{step:6} we test for convergence of the variational approximation.
%We can do this efficiently by comparing estimates from the current and previous
%iteration of the target state at the input points $\bs X$. However, we may
%perform a more robust convergence check and test for correct implementation
%of the algorithm by computing the variational lower bound $\mathcal{L}(q)$ on the model evidence.
%The lower bound should increase at each iteration as the KL-divergence between
%our approximation, $q(\bs Z)$ and the true posterior, $p(\bs Z|\bs c)$,
%decreases. The lower bound for HeatmapBCC is given by:
Due to the approximations described above, we are unable to guarantee an increased variational lower bound for each cycle of the VB algorithm.  We test for convergence of the variational approximation efficiently by comparing the
variational lower bound $\mathcal{L}(q)$ on the model evidence calculated at successive iterations.  The lower bound for HeatmapBCC is given by:
\begin{flalign}
& \mathcal{L}(q) 
 = 
\mathbb{E}_q \!\! \left[\log p\left(\bs c|\bs t,\bs\pi^{(1)}\!\!,..,\bs\pi^{(S)}\right)\right] + \mathbb{E}_q\!\!\left[\log \frac{p(\bs t|\bs\rho)}{q(\bs t)}\right] + \sum_{j=1}^J \!\Bigg\{&& \label{eq:lowerbound}\\
& \left. 
\mathbb{E}_q\!\!\left[\log\!\frac{p\left(\bs f_j | \bs\mu_j, \bs K_j/\varsigma_j\right)}{q(\bs f_j)}\right] \!
 \!+\! \mathbb{E}_q\!\! \left[\log\! \frac{p(\bs\varsigma_j | a_0, b_0)}{q(\bs\varsigma_j)}   \right] \right. 
\left. \!\!
\!+\! \sum_{s=1}^S \mathbb{E}_q\!\! \left[\log\! \frac{p\left(\bs\pi_j^{(s)}|\bs \alpha_{0,j}^{(s)}\right)}{q\left(\bs\pi_j^{(s)}\right)}\right]  \!\!
\right\rbrace  .  &&  \nonumber
\end{flalign}

\paragraph{\bf Predictions:}
%The previous subsections showed how to perform the calculations required in
%steps \ref{step:1} to \ref{step:6} of the VB algorithm. 
%After checking the variational factors for convergence in step \ref{step:6}
Once the algorithm has converged, we predict
 target states, $\bs t^*$ and probabilities $\bs\rho^*$ at output points
$\bs X^*$ by estimating their expected values. 
For a heatmap visualisation, $\bs X^*$ is a set 
of evenly-spaced points on a grid placed over the region of interest. 
We cannot compute the posterior distribution over $\bs\rho^*$
analytically due to the non-linear sigmoid function. 
We therefore estimate the expected values $\mathbb{E}[\bs\rho_j^*]$ by sampling $\bs f_j^*$ from its posterior and mapping the samples through the sigmoid function. 
The multivariate Gaussian posterior of $\bs f_j^*$ has latent mean $\hat{\bs f}^*$ and covariance 
$\bs\Sigma^*$:
\begin{align}
\hat{\bs f}^*_{j} &= \bs\mu^*_j + \bs W_j^*\left(\bs r_j - \sigma(\hat{\bs f}_j) + \bs G(\hat{\bs f}_j - \bs\mu_j )\right) \label{eq:f_mean} \\
\bs\Sigma_j^{*} &= \hat{\bs K}^{**}_j -\bs W_j^* \bs G_j \hat{\bs K}^*_j, \label{eq:f_cov}
\end{align}
where $\bs\mu^*_j$ is the prior mean at the output points, 
$\hat{\bs K}^{**}_j$
% = \{ k_{j,\theta}(\bs x,\bs x') / \mathbb{E}[\varsigma_j], \forall \bs x \in \bs X^*, \forall \bs x\in \bs X^* \}$ 
is the covariance matrix of the output points,  
$\hat{\bs K}^{*}_j$ % = \{ k_{j,\theta}(\bs x,\bs x') / \mathbb{E}[\varsigma_j], \forall \bs x \in \bs X^*, \forall \bs x\in \bs X \}$ 
is the covariance matrix between the input and the output points, 
and $\bs W_j^* = \hat{\bs K}_j^* \bs {G_j}^T \left(\bs G_j \hat{\bs K}_j \bs {G_j}^T + \bs Q_j \right)^{-1}$ is the Kalman gain.
The predictions for output states $\bs t^*$ are the expected probabilities 
$\mathbb{E}\left[ t_{i,j}^*\right] = r^*_{i,j} \propto q(t_i=j,\bs c)$
of each state $j$ at each output point $\bs x_i \in \bs X^*$,
computed using Equation \ref{eq:responsibilities}.
In a multi-class setting, the predictions for each class could be plotted as separate heatmaps.

%% file: sections/experiments.tex
\section{Experiments\label{sec:experiments}} 

We compare the efficacy of our approach with alternative methods on synthetic data and two real datasets.
In the first real-world application we combine crowdsourced annotations of images in the aftermath of a disaster, while in the second we aggregate crowdsourced labels assigned to geo-tagged text messages to predict emergencies in the aftermath of an Earthquake. 
All experiments are binary classification tasks where reports may be negative 
(recorded as $c^{(s)}_{i}=1$)
or positive ($c^{(s)}_{i}=2$). 
In all experiments, we examine the effect of data sparsity using an \emph{incremental train/test procedure}:
\begin{enumerate}
  \item Train all methods on a random subset of reports (initially a small subset)
  \item Predict states $\bs t^*$ at grid points in an area of interest. For HeatmapBCC, we use the predictions $\mathbb{E}[t_{i,j}^*]$ described in Section \ref{sec:vb}
  \item Evaluate predictions using the area under the ROC curve (AUC) 
  %\cite{fawcett_introduction_2006} 
or cross entropy classification error
  \item Increment subset of training labels at random and repeat from step 1.
\end{enumerate}
Specific details vary in each experiment and are described below.
We evaluate HeatmapBCC against the following alternatives: a Kernel density estimator (KDE) \cite{rosenblatt1956remarks,parzen1962estimation}, which is a non-parametric technique
that places a Gaussian kernel at each observation point, 
then normalises the sum of Gaussians over all observations; 
a GP classifier \cite{reece2011determining}, which applies a Bayesian non-parametric approach but assumes reports are equally reliable; IBCC with VB\cite{simpsonlong}, which performs no interpolation between spatial points, but is a state-of-the-art method for combining unreliable crowdsourced classifications; and an ad-hoc combination of IBCC and the GP classifier (\emph{IBCC+GP}), in which the output
classifications of IBCC are used as training labels for the GP classifier. This last method illustrates whether the single VB learning approach of HeatmapBCC is beneficial,
for example, by transferring information between neighbouring data points when learning confusion matrices. 
For the first real dataset, we include additional baselines: SVM with radial basis function kernel; a K-nearest neighbours classifier with $n_{neighbours}=5$ (\emph{NN}); and majority voting (\emph{MV}), which defaults
to the most frequent class label (negative) in locations with no labels .
% reduce repeated text in experiments -- To show the effects of data sparsity, 
% all experiments

\subsection{Synthetic Data}

We ran three experiments with synthetic data to illustrate the behaviour of HeatmapBCC with different types of unreliable reporters. 
For each experiment, we generated $25$ binary ground truth datasets as follows:  
obtain coordinates at all $1600$ points in a $40 \times 40$ grid;
draw latent function values $f_{\bs x}$  from a multivariate Gaussian distribution
 with zero mean and
Mat\'{e}rn$\frac{3}{2}$ covariance with $l=20$ and inverse scale $1.2$; 
apply sigmoid function to obtain state probabilities, $\rho_{\bs x}$;
draw target values, $t_{\bs x}$, at all locations.
\begin{figure}[h]
\begin{center}
\includegraphics[width=0.49\columnwidth,clip=true,trim=18 0 44 27]{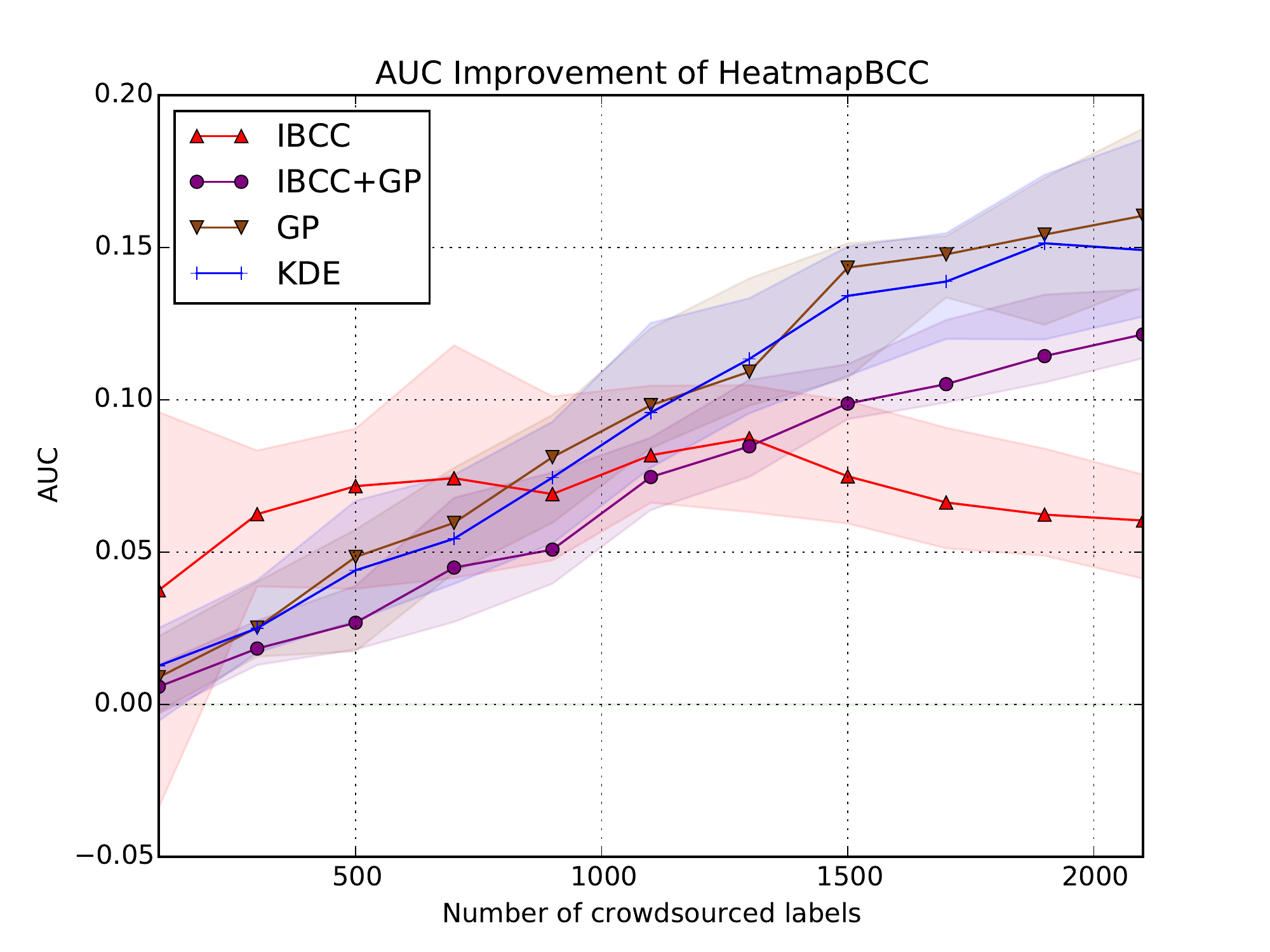}
\includegraphics[width=0.49\columnwidth,clip=true,trim=18 0 44 27]{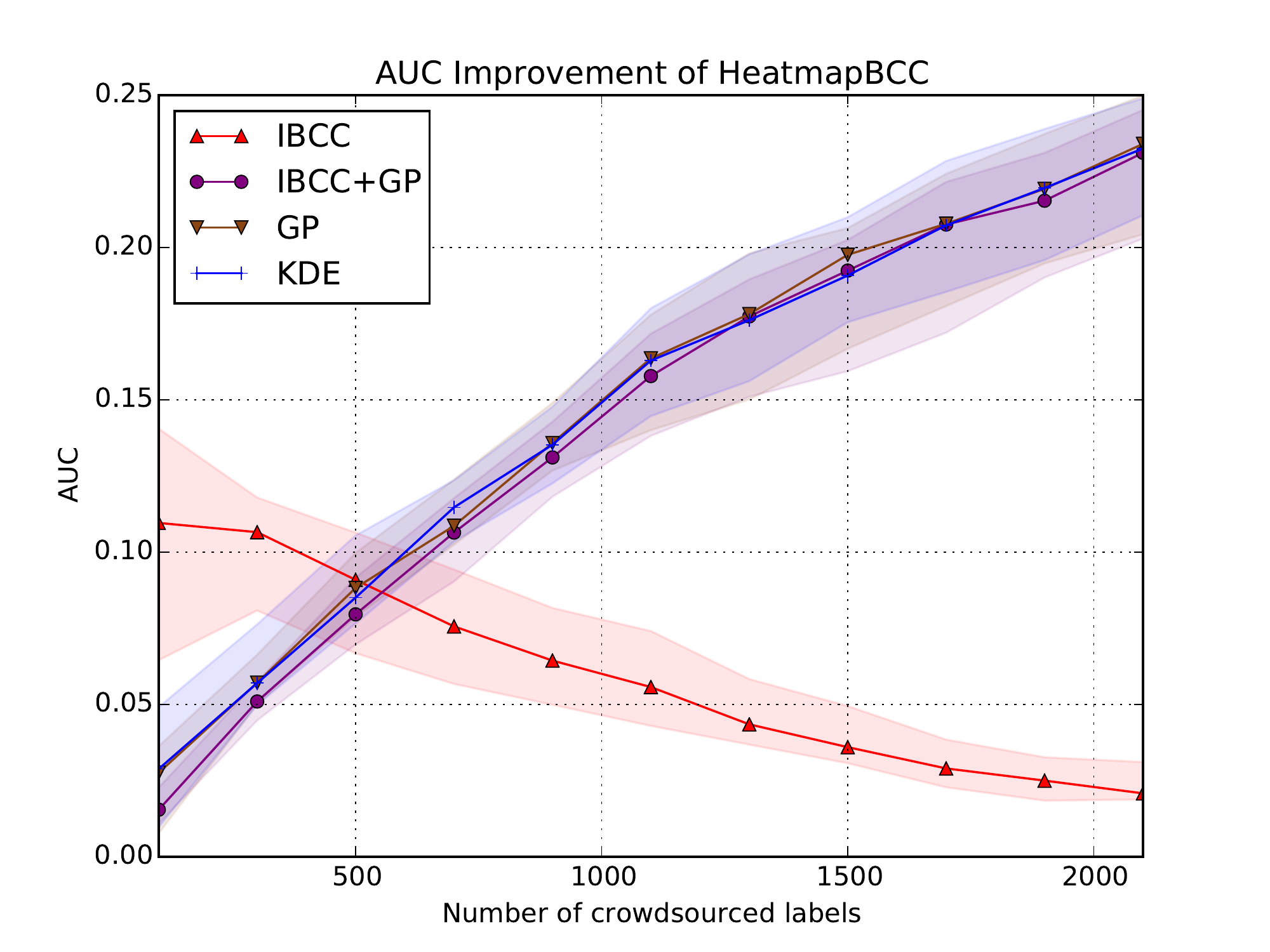}\\
\includegraphics[width=0.49\columnwidth,clip=true,trim=18 10 44 27]{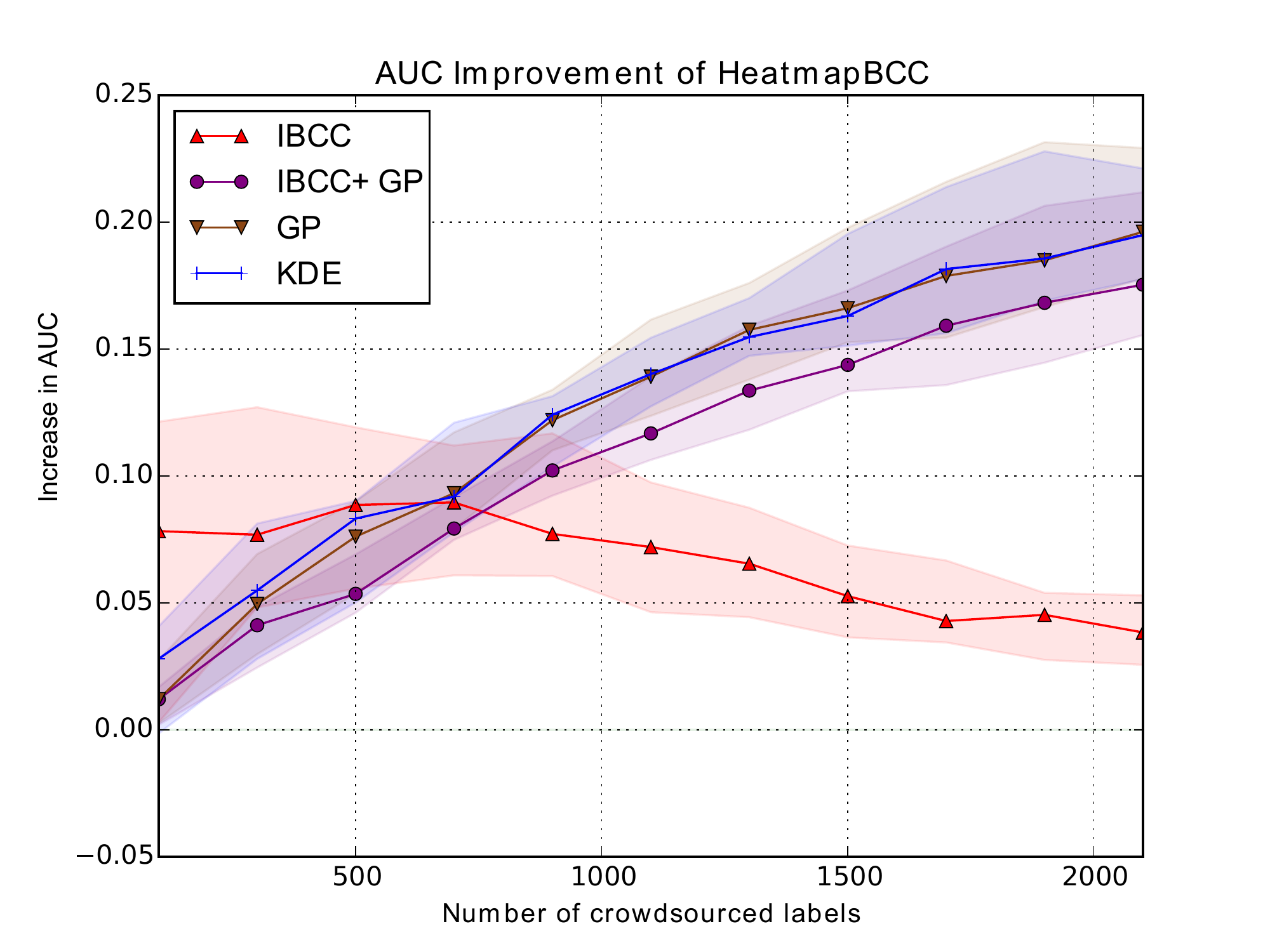}
\includegraphics[width=0.49\columnwidth,clip=true,trim=18 10 44 27]{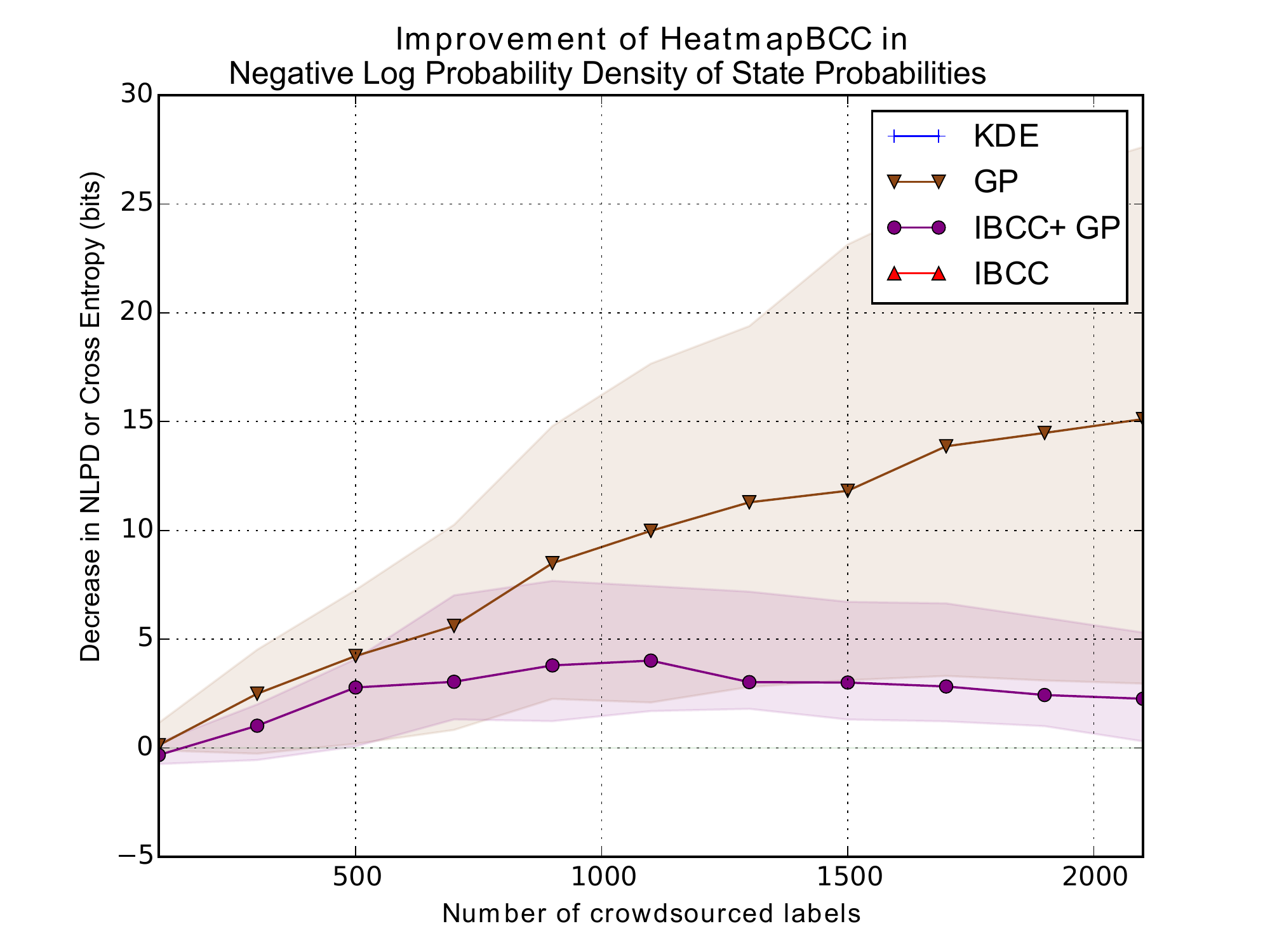}
\end{center}
\caption{\label{synth_noise_aucdiffs}
Synthetic data, \emph{noisy reporters}: median improvement of HeatmapBCC over alternatives
over 25 datasets, 
against number of crowdsourced labels. Shaded areas show inter-quartile range.
 Top-left: AUC, 25\% noisy reporters.
 Top-right: AUC, 50\% noisy reporters.
 Bottom-left: AUC, 75\% noisy reporters.
 Bottom-right: \label{synth_noise_nlpd} NLPD of state probabilities, $\bs\rho$, with 50\% noisy reporters.}
\end{figure}

\emph{Noisy reporters:} the first experiment tests robustness to error-prone annotators. 
For each of the $25$ ground truth datasets, we generated three crowds of $20$ reporters.
In each crowd, we varied the number of \emph{reliable} reporters between $5$, $10$ and $15$,
while the remainder were \emph{noisy} reporters with high random error rates. 
We simulated reliable reporters by drawing confusion matrices, $\bs\pi^{(s)}$, from beta distributions with parameter matrix set to $\bs\alpha^{(s)}_{jj}=10$ along the diagonals and $1$ elsewhere. 
For noisy workers, all parameters were set equally to $\bs\alpha^{(s)}_{jl}=5$.
For each proportion of noisy reporters, we 
selected reporters and grid points at random, and generated $2400$ reports 
by drawing binary labels from the confusion matrices $\bs\pi^{(1)}, ..., \bs\pi^{(20)}$.
We ran the incremental train/test procedure for each crowd with each of the $25$ ground truth datasets.
For HeatmapBCC, GP and IBCC+GP the kernel hyperparameters were set as $l=20$, $a_0=1$, and $b_0=1$. For HeatmapBCC, IBCC and IBCC+GP, we set confusion matrix hyperparameters to $\bs\alpha^{(s)}_{j,j}=2$ along the diagonals and $\bs\alpha^{(s)}_{j,l}=1$ elsewhere, assuming a weak tendency toward correct labels. For IBCC we also set $\nu_0=[1, 1]$.

Figure \ref{synth_noise_aucdiffs} shows the median differences in AUC between HeatmapBCC and the alternative methods for \emph{noisy reporters}. Plotting the difference between methods allows us to see consistent performance differences when AUC varies substantially between runs.
More reliable workers increase the AUC improvement of HeatmapBCC. With all proportions of 
workers, the performance improvements are smaller with very small numbers of labels, except against IBCC, as none of the methods produce a confident model with very sparse data.  
As more labels are gathered, there are more locations with multiple reports, and IBCC is able to make good predictions at those points, thereby reducing the difference in AUC as the number of labels increases. However, for the other three methods, the difference in AUC continues to increase, as they improve more slowly as more labels are received. With more than 700 labels, using the GP to estimate the class labels directly is less effective than using IBCC classifications at points where we have received reports, hence the poorer performance of GP and IBCC+GP.

In Figure \ref{synth_noise_nlpd} we also show the improvement in negative log probability density (NLPD) of state probabilities, $\bs\rho$. We compare HeatmapBCC only against the methods that place a posterior distribution over their estimated state probabilities. As more labels are received, the IBCC+GP method begins to improve slightly, as it is begins to identify the noisy reporters in the crowd. The GP is much slower to improve due to the presence of these noisy labels.
\begin{figure}
\begin{center}
\includegraphics[width=0.497\columnwidth,clip=true,trim=20 13 44 27]{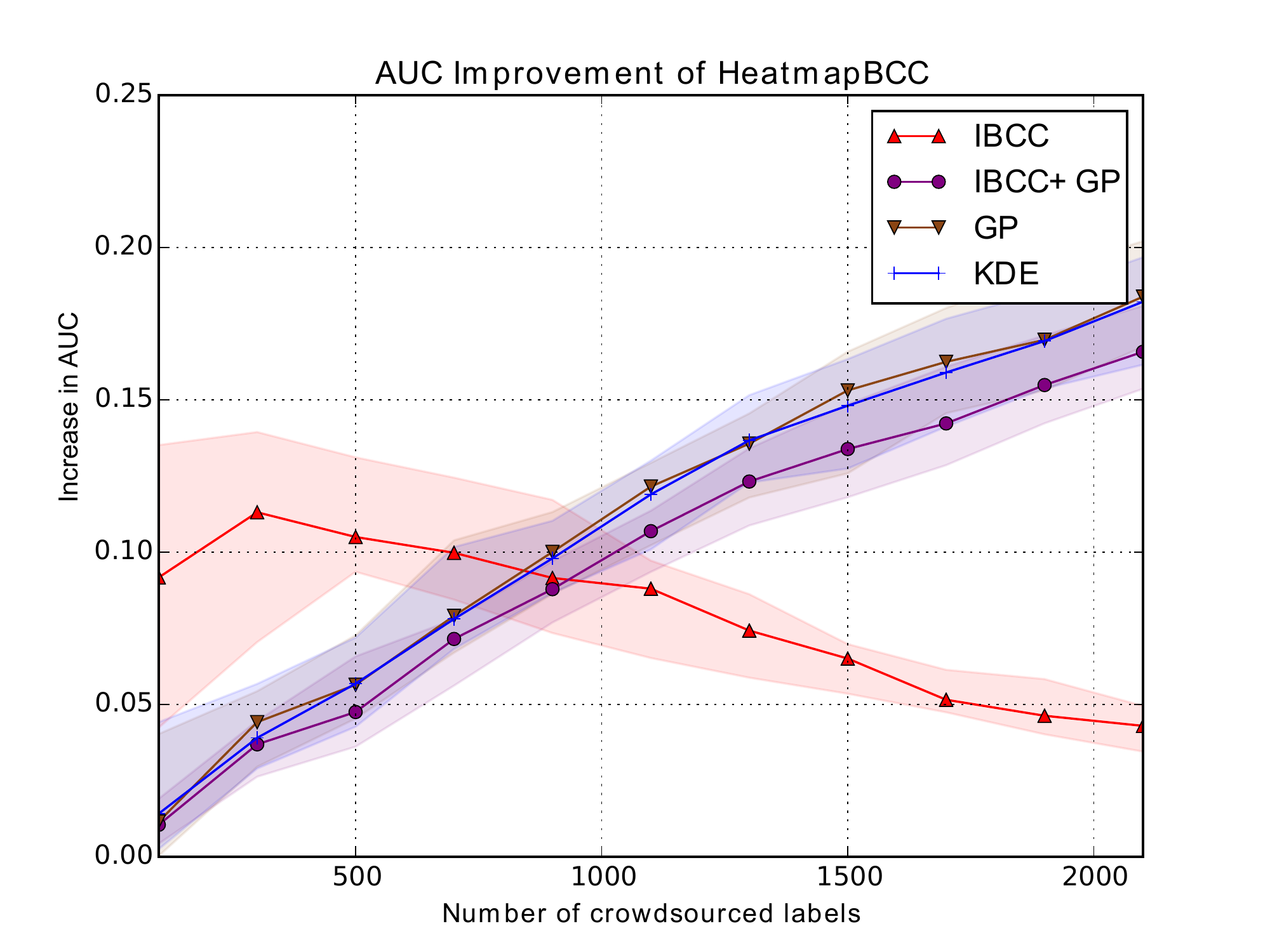}
\includegraphics[width=0.497\columnwidth,clip=true,trim=20 13 44 27]{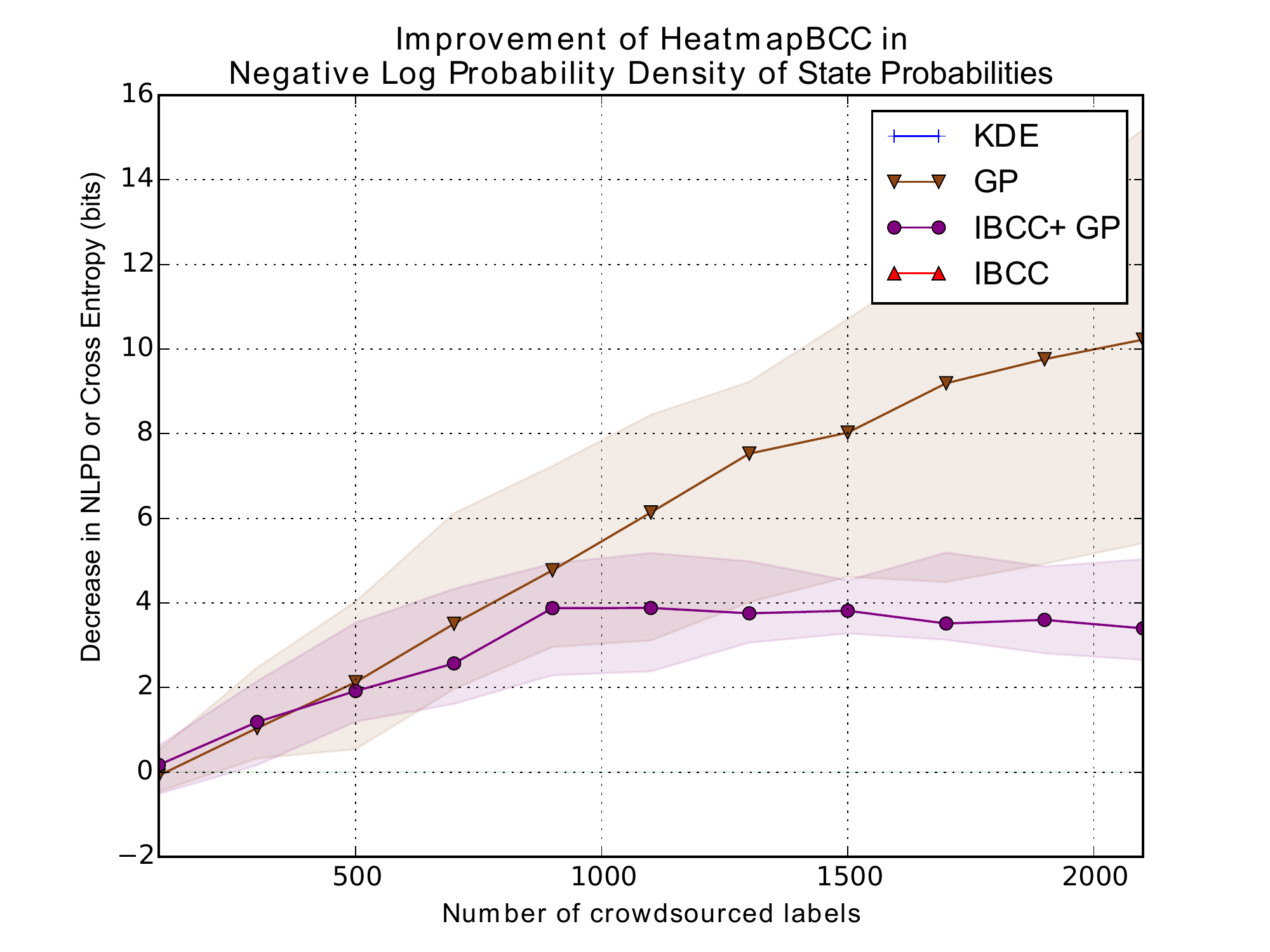}
\end{center}
\caption{\label{synth_bias_auc} Synthetic data, 50\% \emph{biased reporters}:
median improvement of HeatmapBCC compared to alternatives over 25 datasets, against number of crowdsourced labels.
Shaded areas showing the inter-quartile range. 
 Left: AUC. Right: NLPD of state probabilities, $\bs\rho$.}
\end{figure}

\emph{Biased reporters:} the second experiment simulates the scenario where some reporters choose the negative class label overly frequently, e.g. because they fail to observe the positive state when it is present. 
We repeated the procedure used for noisy reporters but replaced the noisy reporters with \emph{biased} reporters generated using the parameter matrix $\bs\alpha^{(s)} = \left[ \begin{smallmatrix}7 & 1\\ 6 & 2\end{smallmatrix}\right]$. 
We observe similar performance improvements to the first experiment with noisy reporters, as shown in 
Figure \ref{synth_bias_auc}, suggesting that HeatmapBCC is also better able to model biased reporters 
from sparse data than rival approaches.
\begin{figure}
\begin{center}
\includegraphics[width=0.497\columnwidth,clip=true,trim=35 552 20 12]{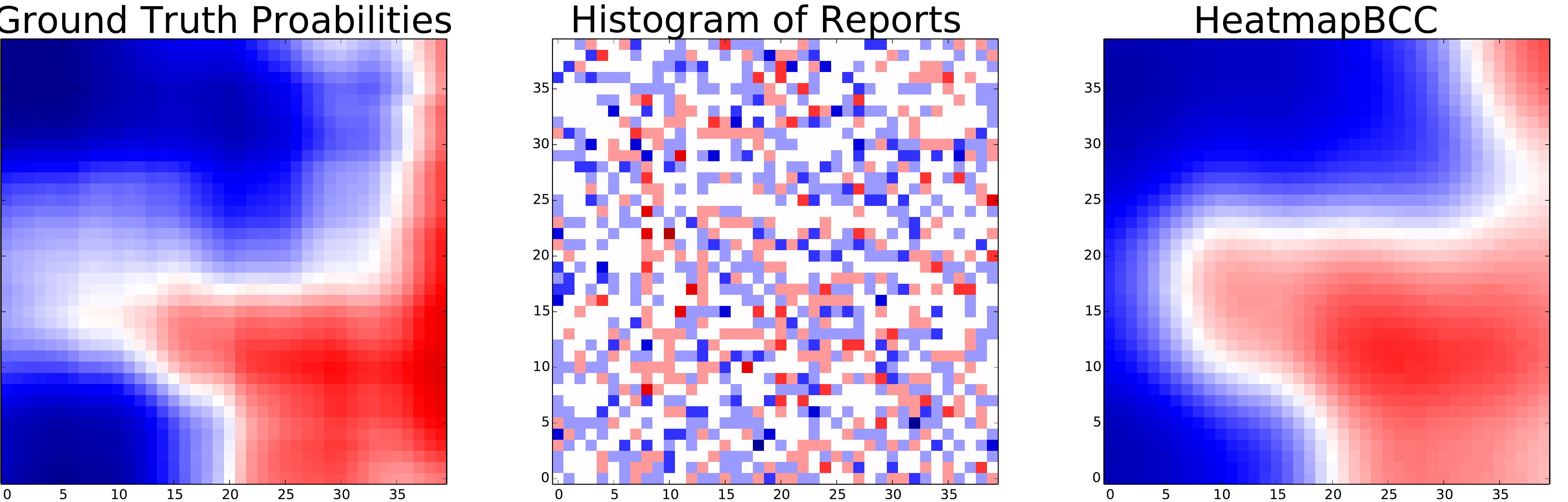}
\includegraphics[width=0.497\columnwidth,clip=true,trim=30 71 35 477]{figures/simulations_bias/casestudy4_main_1500_bias50percent_edited}\\
\includegraphics[width=0.5\columnwidth,clip=true,trim=500 20 500 940]{figures/simulations_bias/casestudy4_main_1500_bias50percent_edited}
\end{center}
\caption{Synthetic data, 50\% \emph{biased reporters}: posterior distributions. Histogram of reports shows the difference between positive and negative label frequencies at each grid square.}\label{casestudy}
\end{figure}
Figure \ref{casestudy} shows an example of the posterior distributions over $t_{\bs x}$ produced by each method when trained on $1500$ random labels from a simulated crowd with $50\%$ \emph{biased reporters}. 
We can see that the ground truth appears most similar to the HeatmapBCC estimates, while IBCC is unable to perform any smoothing.

\begin{figure}
\begin{center}
\includegraphics[width=0.497\columnwidth,clip=false,trim=18  15 44 27]{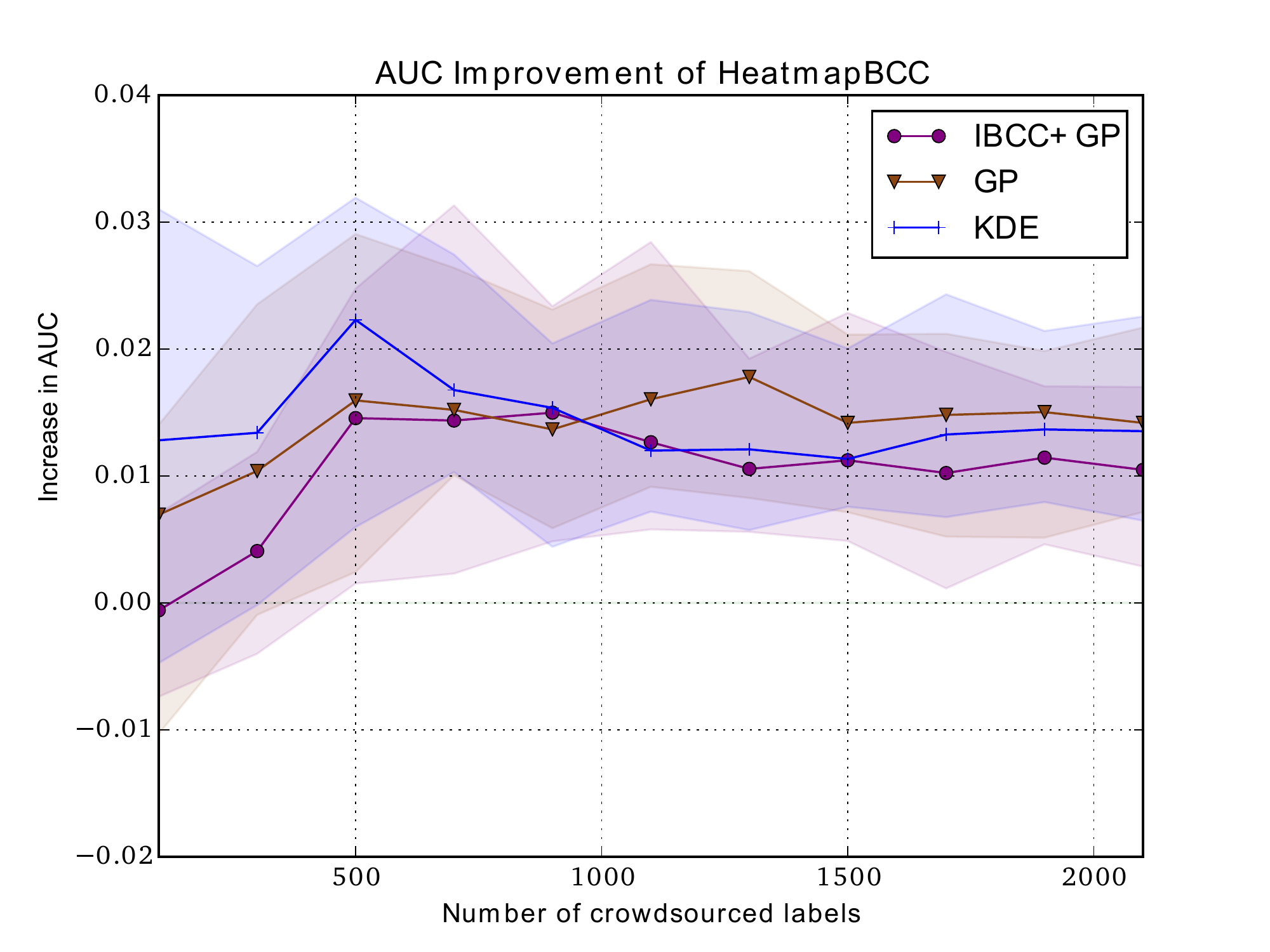}
\includegraphics[width=0.497\columnwidth,clip=false,trim=18 15 44 27]{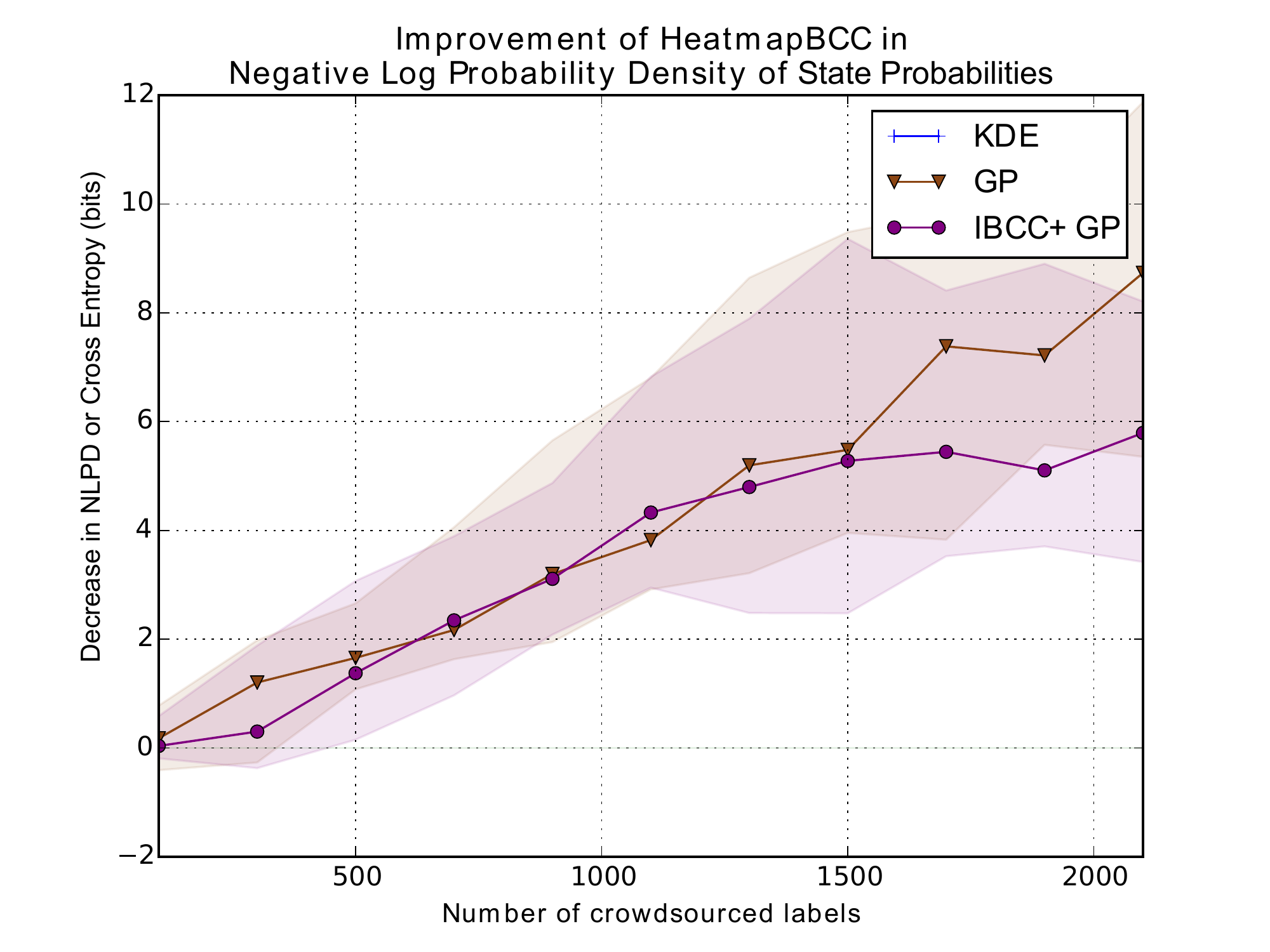}
\end{center}
\caption{\label{synth_nogrid_auc} Synthetic data, 50\% \emph{noisy reporters,
continuous report locations}. Median improvement of HeatmapBCC compared to alternatives over 25 datasets, against number of crowdsourced labels.
Shaded areas showing the inter-quartile range. 
Left: AUC. Right: \label{synth_nogrid_nlpd} NLPD of state probabilities, $\bs\rho$. }
\end{figure}

\emph{Continuous report locations}: in the previous experiments we drew reports from discrete grid points
so that multiple reporters produced noisy labels for the same target, $t_{\bs x}$. 
The third experiment tests the behaviour of our model with reports drawn from continuous locations, with 50\% noisy reporters drawn as in the first experiment.
In this case, our model receives only one report for each object $t_{\bs x}$ at the input locations $\bs X$. Figure \ref{synth_nogrid_auc} shows that the difference in AUC between HeatmapBCC and other methods is significantly reduced, although still positive. This may be because we are reliant on $\rho$ to make classifications, since we have not observed any reports for the exact test locations $\bs X^*$. 
If $\rho_{\bs x}$ is close to $0.5$, the prediction for class label $\bs x$ is uncertain. 
However, the improvement in NLPD of the state probabilities $\bs\rho$ is less affected 
by using continuous locations, 
as seen by comparing Figure \ref{synth_noise_nlpd} with Figure \ref{synth_nogrid_nlpd}, 
suggesting that HeatmapBCC remains advantageous when there is only one report at each training location. 
In practice, reports at neighbouring locations may be intended to refer to the same $t_{\bs x}$, so if reports are treated as all relating to separate objects, they could bias the state probabilities. Grouping reports into discrete grid squares avoids this problem and means we obtain a state classification for each square in the heatmap. We therefore continue to use discrete grid locations in our real-world experiments.
%Therefore the choice of whether to group reports into discrete grid squares, or to treat continuous reports as relating to separate objects is application specific.

\subsection{Crowdsourced Labels of Satellite Images}

We obtained a set of 5,477 crowdsourced labels from a trial run of the Zooniverse Planetary Response Network project\footnote{\url{http://www.planetaryresponsenetwork.com/beta/}}. In this application,
volunteers labelled satellite images showing damage to Tacloban, Philippines, after Typhoon Haiyan/Yolanda. The volunteers' task was to mark features such as damaged buildings, blocked roads and floods.
For this experiment, we first divided the area into a $132 \times 92$ grid. 
The goal was then to combine crowdsourced labels to classify grid squares according to whether they contain buildings with major damage or not. 
We treated cases where a user observed an image but did not mark any features as a set of multiple negative labels, one for each of the grid squares covered by the image. 
Our dataset contained 1,641 labels marking buildings with major structural damage, and 1,245 negative labels. 
Although this dataset does not contain ground truth annotations, it contains enough crowdsourced annotations that we can confidently determine labels for most of the region of interest using all data. The aim is to test whether our approach can replicate these results using only a subset of crowdsourced labels, thereby reducing the workload of the crowd by allowing for sparser annotations. 
We therefore defined gold-standard labels by running IBCC on the complete set of crowdsourced labels,
and then extracting the IBCC posterior probabilities for $572$ data points with $\geq 3$ crowdsourced labels where the posterior of the most probable class $\geq 0.9$.
The IBCC hyperparameters were set to $\bs\alpha^{(s)}_{0,j,j}=2$ along the diagonals, $\bs\alpha_{0,j,l}^{(s)}=1$ elsewhere, and $\nu_0=[100, 100]$.

%Since these are confident classifications obtained using multiple labels per data point, we treat these as reliable labels. We now wish to determine whether HeatmapBCC can match these classifications with only a subset of the crowdsourced labels, thereby showing that HeatmapBCC can reduce the workload of the crowd. 

We ran our incremental train/test procedure 20 times with initial subsets of 178 random labels. 
Each of these 20 repeats required approximately 45 minutes runtime on an Intel i7 desktop computer.
The length-scales $l$ for HeatmapBCC, GP and IBCC+GP were optimised at each iteration using maximum likelihood II by maximising the variational lower bound on the log likelihood (Equation \ref{eq:lowerbound}), as described in \cite{rasmussen_gaussian_2006}. 
The inverse scale hyperparameters were set to $a_0=0.5$ and $b_0=5$, and the other hyperparameters were set as for gold label generation. We did not find a significant difference when varying diagonal confusion matrix values $\bs\alpha^{(s)}_{j,j}=2$ from $2$ to $20$.
% not known in advance, and the complex non-conjugate relationship between $l$ and the rest of the model means that we were not able to marginalise $l$ as part of our variational inference algorithm. We therefore 
%marginalise the length-scale approximately using  importance sampling from a gamma prior $p(l | a_l, b_l)$ with %shape $a_l=2$ and scale $1/b_l = 22$.

% In this scenario, the length-scale is not known in advance, so we , then ran all methods with the same set of sample length-scales from this prior. We then weighted the predictions according to $\mathcal{L} + \log p(l | a_l, b_l)$. In future, physical knowledge or data from previous, similar scenarios can also be used to choose a suitable length-scale. For the confusion matrix hyper-parameters, we set $\alpha^{(s)}_{j,j}=5$ along the diagonals, and $\alpha^{(s)}_{j,\neg j}=1$ on the off-diagonals. We set $a_0=0.5$ and $b_0=5$, and use a zero mean function $m_0$. 

\begin{figure}
\begin{center}
\includegraphics[width=0.497\columnwidth,clip=true,trim=25 13 45 27]{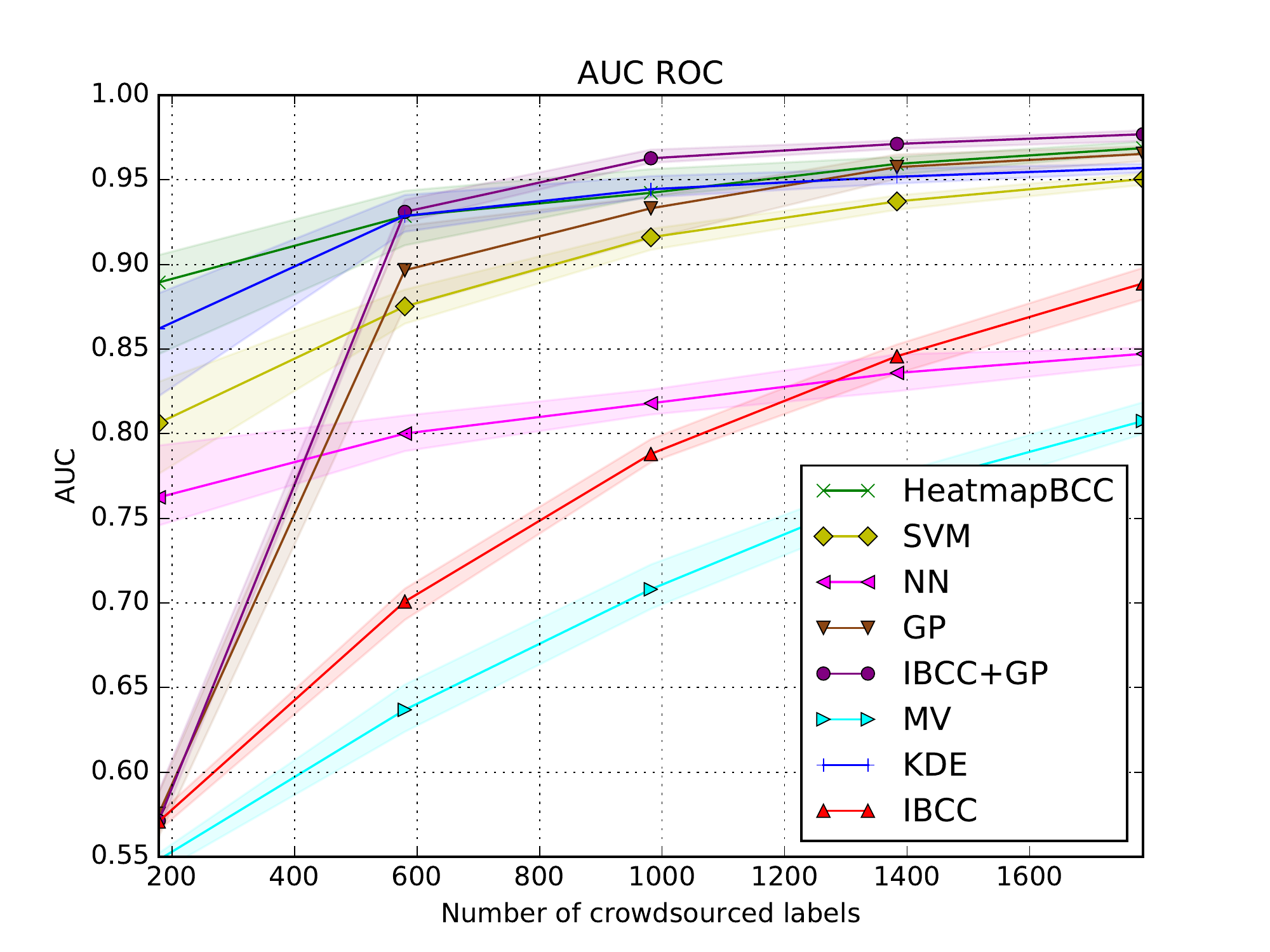}
\includegraphics[width=0.497\columnwidth,clip=true,trim=25 13 45 27]{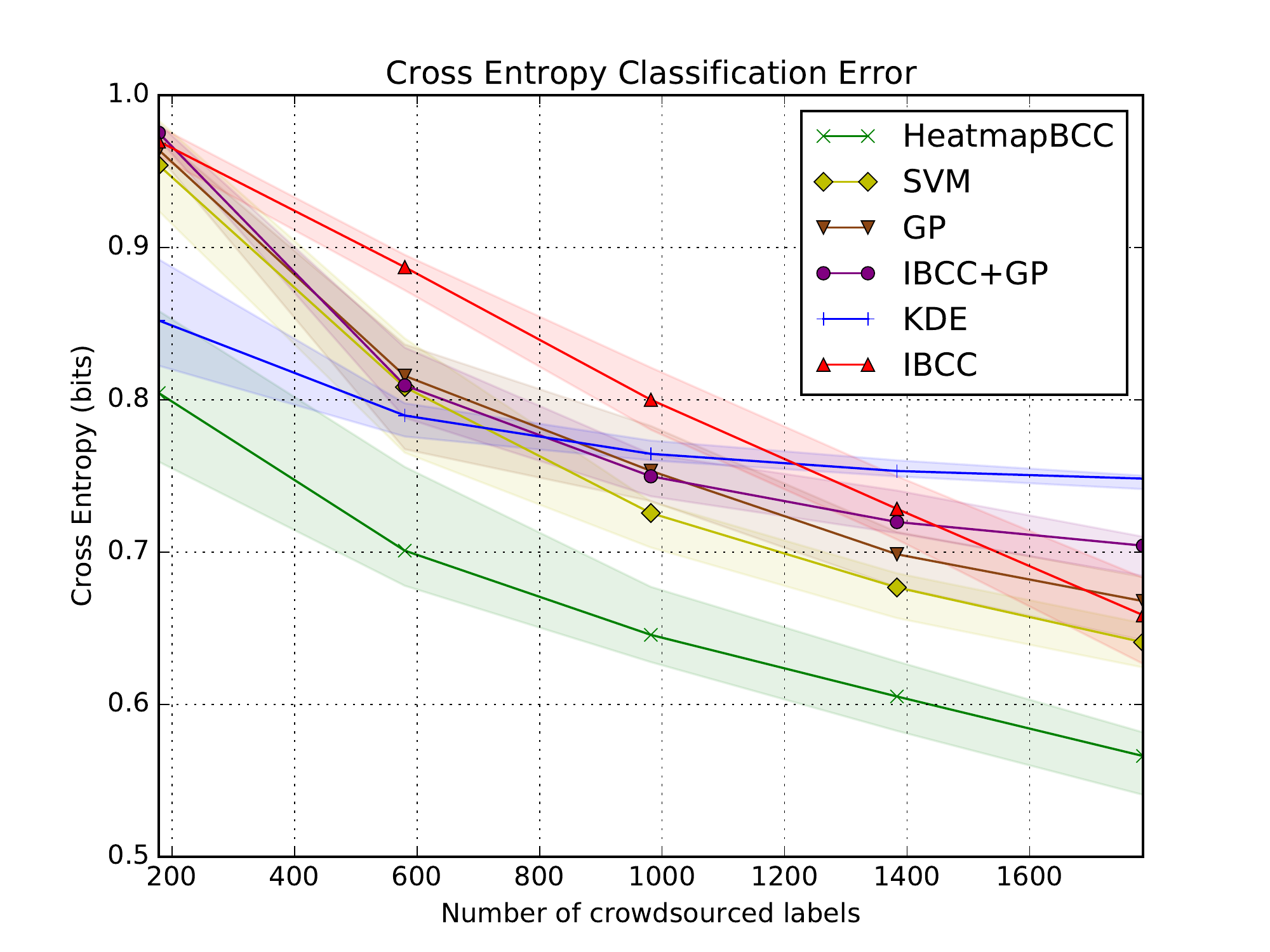}
\end{center}
\caption{\label{prn_sd3_auc} Planetary Response Network, major structural damage data.
Median values over 20 repeats against the number of randomly selected crowdsourced labels.
Shaded areas show the inter-quartile range.
Left: AUC. Right: \label{prn_sd3_ce} cross entropy error.}
\end{figure}
In Figure \ref{prn_sd3_auc} (left) we can see how AUC varies as more labels are introduced, with HeatmapBCC, GP and IBCC+GP converging close to our gold-standard solution. HeatmapBCC performs best initially, potentially because it can learn a more suitable length-scale with less data than GP and IBCC+GP. 
SVM outperforms GP and IBCC+GP with $178$ labels, but is outperformed when more labels are provided.
Majority voting, nearest neighbour and IBCC produce much lower AUCs than the other approaches. 
The benefits of HeatmapBCC can be more clearly seen in Figure \ref{prn_sd3_ce} (right), which shows a substantial reduction in cross entropy classification error compared to alternative methods, indicating that HeatmapBCC produces better probability estimates.

\subsection{Haiti Earthquake Text Messages}

Here we aggregate text reports written by members of the public after the Haiti 2010 Earthquake.
The dataset we use was collected and labelled by Ushahidi\cite{morrow2011independent}. 
We have selected 2,723 geo-tagged reports that were sent mainly by SMS and were categorised by Ushahidi volunteers.
The category labels describe the type of situation that is reported, such as ``medical emergency" 
or ``collapsed building". 
In this experiment, we aim to predict a binary class label, "emergency" or "no emergency"
by combining all reports. 
We model each category as a different information source; if a category label is present for a particular message, we observe a value of $1$ from that information source at the message's geo-location.
% HeatmapBCC can accurately aggregate these sparsely-distributed reports into a situation awareness heatmap showing likely emergencies.% accounting for the informativeness of each category label. 
%The heatmap should indicate where likely emergencies are taking place that first responders should attend, such as medical emergencies or trapped people. 
This application differs from the satellite labelling task because many of the reports do not explicitly report emergencies and may be irrelevant.
%unlike workers in a crowdsourcing environment, members of the public have different motivations and information available when making a report, and many of the category labels assigned to reports may not be relevant to the task of locating emergencies. 
%A goal of our method is to predict the state at locations where no reports have been received, by interpolating between neighbouring locations. 
%Given an exhaustive dataset with several reports per data point, the standard IBCC model with no interpolation has been shown to give accurate results \cite{simpsonlong}. Therefore, 
In the absence of
ground truth data, we establish a gold-standard test set by training IBCC on all 2723 reports, placed into 675 discrete locations on a $100 \times 100$ grid. Each grid square has approximately 4 reports. We set IBCC hyper-parameters to $\bs\alpha^{(s)}_{0,j,j}=100$ along the diagonals, $\bs\alpha^{(s)}_{0,j,l}=1$ elsewhere, and $\bs\nu_0=[2000, 1000]$.

Since the Ushahidi data set contains only reports of emergencies, and does not contain reports stating that no emergency is taking place, we cannot learn the length-scale $l$ from this data, and must rely on background knowledge. 
We therefore select another dataset from the Haiti 2010 Earthquake, which has gold standard labels, namely the building damage assessment provided by UNOSAT \cite{corbane2011comprehensive}. We expect this data to have 
a similar length-scale because the underlying cause of both the building damages and medical emergencies was an earthquake affecting built-up areas where people were present. 
We estimated $l$ using maximum likelihood II optimisation, giving an optimal value of $l=16$ grid squares. We then transferred this point estimate to the model of the Ushahidi data.
Our experiment repeated the incremental train/test procedure 20 times with hyperparameters set to $a_0=1500$, 
$b_0=1500$, $\bs\alpha^{(s)}_{0,j,j}=100$ along the diagonals, $\bs\alpha^{(s)}_{0,j,l}=1$ elsewhere, and $\bs\nu_0=[2000, 1000]$.

\begin{figure}
\begin{center}
\includegraphics[width=0.495\columnwidth,clip=true,trim=25 13 48 27]{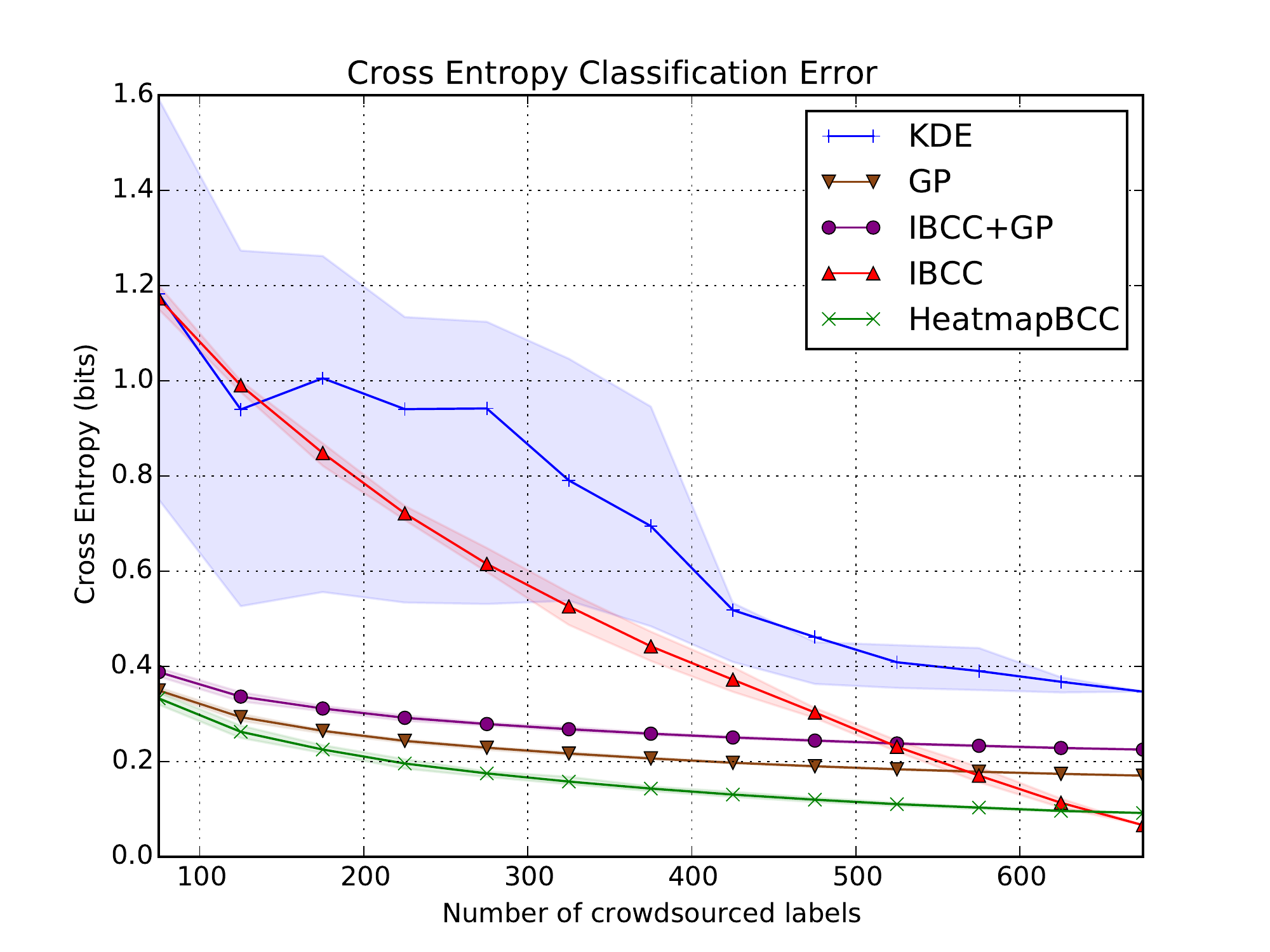}
\includegraphics[width=0.499\columnwidth,clip=true,trim=1mm 0 4mm 0mm]{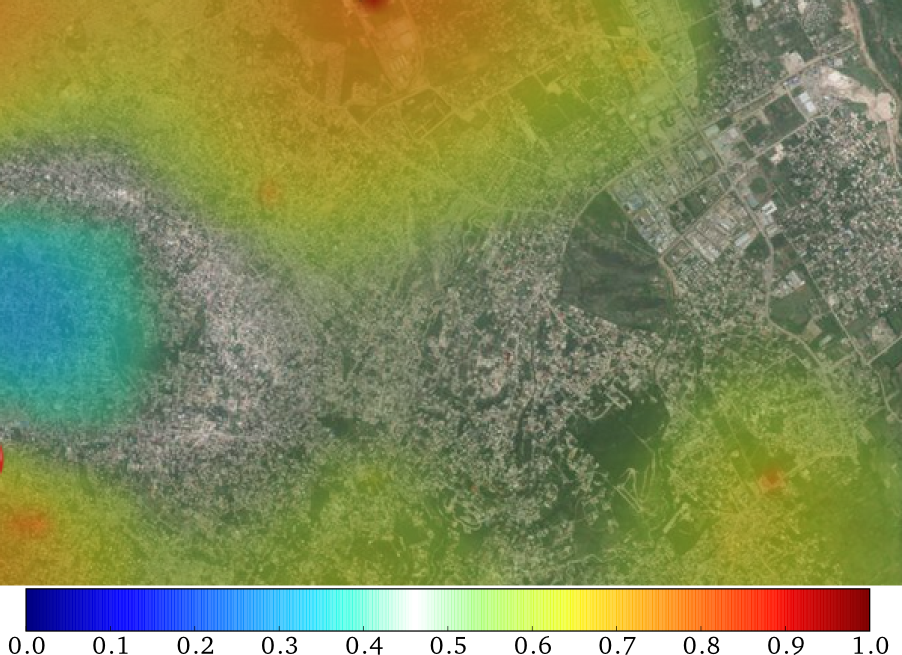}
\end{center}
\caption{Haiti text messages. Left: cross entropy error against the number of randomly selected crowdsourced labels. Lines show the median over 25 repeats, with shaded areas showing the inter-quartile range. Gold standard defined by running IBCC with 675 labels using a $100 \times 100$ grid.
Right: \label{crowdscanner} heatmap of emergencies for part of Port-au-Prince after
the 2010 Earthquake, showing high probability (dark orange) to low probability (blue).
%Targets for UAVs are marked identified by red `?' icons.
}
\label{haiti_cross_entropy} 
\end{figure}
Figure \ref{haiti_cross_entropy} shows that HeatmapBCC 
is able to achieve low error rates when the reports are sparse.
The IBCC and HeatmapBCC results do not quite converge due to the effect of interpolation performed by HeatmapBCC, which can still affect the results with several reports per grid square. The gold-standard predictions from IBCC also contain some uncertainty, so cross entropy does not reach zero, even with all labels. The GP alone is unable to determine the different reliability levels of each report type, so while it is able to interpolate between sparse reports, HeatmapBCC and IBCC detect the reliable data and produce different predictions when more labels are supplied. In summary, HeatmapBCC produces predictions with 439 labels (65\%) that has an AUC within 0.1 of the gold standard predictions produced using all 675 labels, and reduces cross entropy to 0.1 bits with 400 labels (59\%), showing that it is effective at predicting emergency states with reduced numbers of Ushahidi reports. Using an Intel i7 laptop, the HeatmapBCC inference over 675 labels required approximately one minute.

We use HeatmapBCC to visualise emergencies in Port-au-Prince, Haiti after the 2010 earthquake, by plotting the posterior class probabilities as the heatmap shown in Figure \ref{crowdscanner}. 
Our example shows how HeatmapBCC can combine reports from trusted sources with crowdsourced information. 
The blue area shows a negative report from a simulated first responder, with confusion matrix hyperparameters 
set to $\bs\alpha^{(s)}_{0,j,j}=450$ along the diagonals, so that the negative
report was highly trusted and had a stronger effect than the many surrounding positive reports.
Uncertainty in the latent function $f_j$ can be used to identify regions where information is lacking
and further reconnaisance is necessary.
Probabilistic heatmaps therefore offer a powerful tool for situation awareness and planning in disaster response.

%% file: sections/discussion.tex
\section{Conclusions\label{sec:dicussion}} 

In this paper we presented a novel Bayesian approach to aggregating unreliable discrete observations from different sources to classify the state across a region of space or time. We showed how this method can be used to combine noisy, biased and sparse reports and interpolate between them to produce probabilistic spatial heatmaps for applications such as situation awareness.
Our experiments demonstrated the advantages of integrating a confusion matrix model to capture the unreliability of different information sources with sharing information between sparse report locations using Gaussian processes. 
In future work we intend to improve scalability of the GP using stochastic variational inference \cite{hensman_scalable_2015}
and investigate clustering confusion matrices using a hierarchical prior, as per \cite{moreno_bayesian_2015,venanzi2014community}, which may improve the ability to learn confusion matrices when 
data for individual information sources is sparse.
%In this paper we focus on the combination of confusion matrices and GPs, and leave
%refining the confusion matrix prior to future work.
%In comparison to using a Gaussian Process classifier, our approach detects the reliability of different sources and adjusts the posterior distribution according to their informativeness. 
%The relationships between neighbouring points also help HeatmapBCC to learn the confusion matrices for each information source, resulting in better performance than when IBCC first and feeding its predictions into a GP density estimator.  